\documentclass[10pt,twocolumn,letterpaper]{article}

\usepackage{iccv}
\usepackage{times}
\usepackage{epsfig}
\usepackage{graphicx}
\usepackage{amsmath}
\usepackage{amssymb}
\usepackage[accsupp]{axessibility}

\usepackage[utf8]{inputenc} 
\usepackage[T1]{fontenc}    
\usepackage[spaces,hyphens]{url}     
\usepackage{booktabs}       
\usepackage{amsfonts}       
\usepackage{nicefrac}       
\usepackage{microtype}      
\usepackage{xcolor}         
\usepackage{algorithm}
\usepackage{algpseudocode}
\usepackage{fp}
\usepackage{wrapfig,lipsum,booktabs,epstopdf}
\usepackage{lipsum}
\usepackage{rotating}
\usepackage{multirow}
\usepackage{algpseudocode}
\usepackage{pifont}
\newcommand{\cmark}{\ding{51}}
\newcommand{\xmark}{\ding{55}}
\usepackage{array, makecell} 
\usepackage{xcolor,colortbl}
\usepackage{comment}
\usepackage{kantlipsum}
\usepackage{tabularx}
\usepackage{booktabs}       
\usepackage{amsmath}
\usepackage{diagbox}
\usepackage{fancyhdr}
\usepackage{float}
\usepackage{amsfonts} 
\usepackage{caption}
\usepackage{graphicx}
\usepackage{subcaption}
\usepackage{etoolbox}
\usepackage{enumitem}
\usepackage{adjustbox}
\setlist{topsep=0pt,noitemsep,topsep=0pt,parsep=0pt,partopsep=0pt}
\definecolor{bgreen}{rgb}{0.0, 0.5, 0.0}
\definecolor{deepskyblue}{rgb}{0.0, 0.75, 1.0}

\usepackage[pagebackref=true,breaklinks=true,letterpaper=true,colorlinks,bookmarks=false]{hyperref}

\iccvfinalcopy 

\def\iccvPaperID{9199} 

\ificcvfinal\fi

\begin{document}

\newcommand{\papername}{Holistic, Reliable and Scalable Benchmark \\ for Text-to-Image Models}
\newcommand{\papernameAbbrev}{HRS-Bench}
\newcommand{\faizan}[1]{\textcolor{blue}{Faizan:\ #1}}
\newcommand{\shen}[1]{~\textcolor{pink}{Shen:\ #1}}
\newcommand{\pengzhan}[1]{~\textcolor{red}{Pengzhan:\ #1}}
\newcommand{\me}[1]{~\textcolor{cyan}{elhoseiny:\ #1}}

\title{{\papernameAbbrev}: {\papername}}

\author{
Eslam Mohamed Bakr$^{1}$, 
Pengzhan Sun$^{2}$ $^*$,
Xiaoqian Shen $^{1}$ $^*$, \\
Faizan Farooq Khan $^{1}$ $^*$, 
Li Erran Li$^{3}$, 
Mohamed Elhoseiny$^{1}$ \\
\texttt{\{eslam.abdelrahman, xiaoqian.shen, faizan.khan,} \\ \texttt{mohamed.elhoseiny\}@kaust.edu.sa} \\
\texttt{pengzhan@comp.nus.edu.sg, lilimam@amazon.com} \\
$^{1}$King Abdullah University of Science and Technology (KAUST) \hspace{0.1cm} \\
$^{2}$National University of Singapore \hspace{0.1cm}
$^{3}$AWS AI, Amazon  \hspace{0.1cm}
}

\maketitle

\def\thefootnote{*}\footnotetext{Equal Contribution}
\ificcvfinal\thispagestyle{empty}\fi

\begin{abstract}
    In recent years, Text-to-Image (T2I) models have been extensively studied, especially with the emergence of diffusion models that achieve state-of-the-art results on T2I synthesis tasks. 
    However, existing benchmarks heavily rely on subjective human evaluation, limiting their ability to holistically assess the model's capabilities. 
    Furthermore, there is a significant gap between efforts in developing new T2I architectures and those in evaluation. 
    To address this, we introduce {\papernameAbbrev}, a concrete evaluation benchmark for T2I models that is \textbf{H}olistic, \textbf{R}eliable, and \textbf{S}calable. 
    Unlike existing benchmarks that focus on limited aspects, {\papernameAbbrev} measures 13 skills that can be categorized into five major categories: accuracy, robustness, generalization, fairness, and bias. 
    In addition, {\papernameAbbrev} covers 50 scenarios, including fashion, animals, transportation, food, and clothes. We evaluate nine recent large-scale T2I models using metrics that cover a wide range of skills. A human evaluation aligned with 95\% of our evaluations on average was conducted to probe the effectiveness of {\papernameAbbrev}. 
    Our experiments demonstrate that existing models often struggle to generate images with the desired count of objects, visual text, or grounded emotions. We hope that our benchmark help ease future text-to-image generation research.
    The code and data are available at \href{https://eslambakr.github.io/hrsbench.github.io/}{https://eslambakr.github.io/hrsbench.github.io/}.
\end{abstract}

\vspace{-0.8cm}


\section{Introduction}
\label{sec_introduction}

\begin{figure}
\begin{center}
\includegraphics[width=0.99\linewidth]{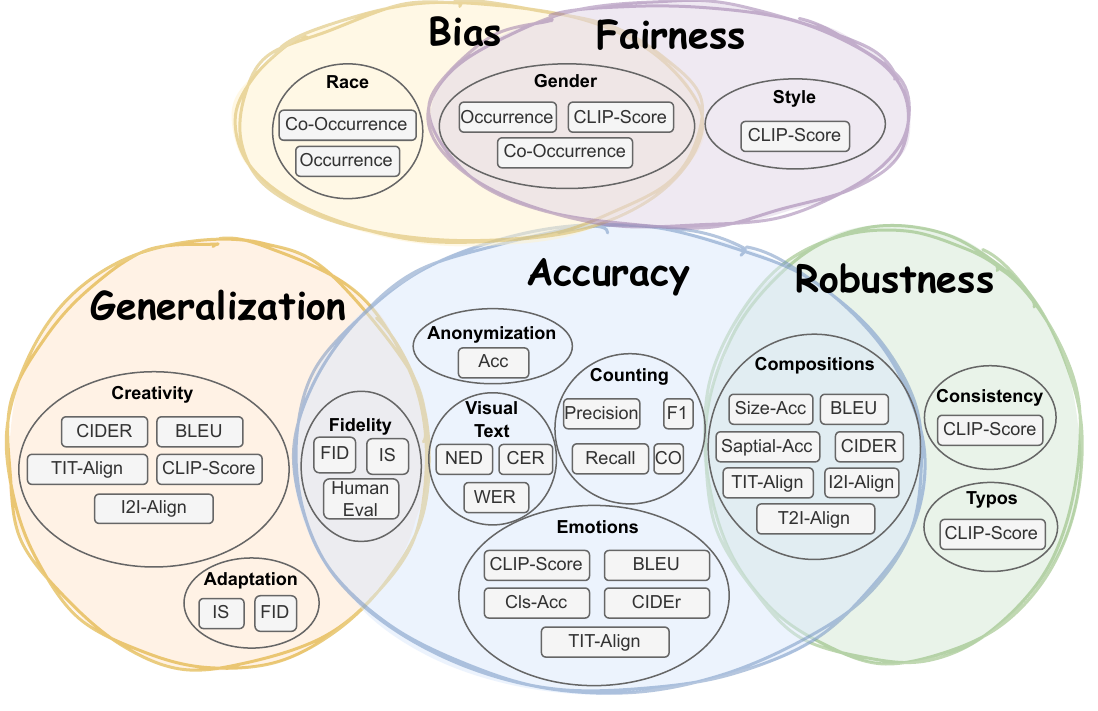}
\end{center}
\caption{An overview of our proposed benchmark, {\papernameAbbrev}, measures 13 skills which could be grouped into five major categories; accuracy, robustness, generalization, fairness, and bias.}
\label{fig_skills_metric}
\vspace{-0.4cm}
\end{figure}

\tabcolsep=0.08cm
\begin{table*}
\centering
\caption{Comparison of text-to-image benchmarks in terms of:
 1) The number of evaluated models. 2) The number of the covered skills. 3) The number of utilized metrics. 4) The evaluation type, whether human or metric based or both. 5) The number of prompts. 6) The prompt generation type, whether a template, human-based or both. 7) Whether there are different hardness levels are included. 8) Number of annotators contributed to the evaluation.}
 \scalebox{0.7}{
\begin{tabular}{ccccccccccc}
    \toprule[1.5pt]
    \multirow{2}{*}{\textbf{Method}} & \- & \- & \- & \multicolumn{2}{c}{\textbf{Eval Type}} &  \- & \multicolumn{2}{c}{\textbf{Prompt Type}} \\ 
    \- &  \textbf{\# Models} &  \textbf{\# Skills} & \textbf{\# Metrics}  &  \textbf{Human} & \textbf{Auto} & \textbf{\# Prompts} & \textbf{Template} & \textbf{Human} & \makecell{\textbf{Hardness}\\\textbf{Levels}} & \textbf{\# Annotators}\\
   \midrule[0.75pt]
    \textbf{DrawBench} \cite{saharia2022photorealistic} & 5 & 4 & 0 & \cmark & \xmark & 200  & \xmark & \cmark & \xmark & 25\\
    \textbf{DALL-EVAL} \cite{cho2022dall} & 4 & 5 & 3 & \cmark & \xmark & 7330 & \cmark & \xmark & \xmark & 6\\
    \textbf{HE-T2I} \cite{petsiuk2022human} & 2 & 3 & 0 & \cmark & \xmark & 90 & \xmark & \cmark & \cmark & 20\\
    \textbf{TISE} \cite{dinh2022tise} & 7 & 3 & 5 & \xmark & \cmark & N/A & \xmark & \cmark & \xmark & N/A\\
    \textbf{{\papernameAbbrev} (Ours)} & \textbf{9} & \textbf{13} & \textbf{17} & \textbf{\cmark} & \textbf{\cmark} & \textbf{45000} & \textbf{\cmark} & \textbf{\cmark} & \textbf{\cmark} & \textbf{1000}\\

  \bottomrule[1.5pt]
\end{tabular}
}
\label{tab_benchmarks_comparision}
\end{table*}

Text-to-Image Synthesis (T2I), one of the essential multi-modal tasks, witnessed remarkable progress starting from conditional GANs \cite{reed2016generative, nguyen2017plug, zhang2017stackgan, hong2018inferring, zhu2019dm}, which are shown to work on simple datasets \cite{nilsback2008automated, wah2011caltech, yu2015lsun, lin2014microsoft}, to recently diffusion models \cite{ding2021cogview, saharia2022photorealistic, dhariwal2021diffusion, chang2023muse, rampas2022fast, ramesh2022hierarchical, zhou2022towards, nichol2021glide, rombach2022high}, which are trained on large-scale datasets, e.g., LAION \cite{schuhmann2021laion, schuhmann2022laion}.

Despite the rapid progress, the existing models face several challenges, e.g., they cannot generate complex scenes with the desired objects and relationship composition \cite{feng2022training, liu2022compositional}.
Furthermore, assessing the T2I models should include more than just fidelity, e.g., the ability to compose multiple objects and generate emotionally grounded or creative images.
Therefore, some recent efforts are focusing on improving the existing metrics \cite{dinh2022tise} or proposing new metrics that cover new aspects, such as bias \cite{zhang2023auditing}, compositions \cite{liu2022compositional, feng2022training, park2021benchmark}.
Moreover, some other works propose new benchmarks, summarized in Table \ref{tab_benchmarks_comparision}, that assess different aspects, e.g., counting \cite{saharia2022photorealistic, cho2022dall, petsiuk2022human}, social-bias \cite{cho2022dall}, and object fidelity \cite{dinh2022tise, petsiuk2022human}.
Even with various benchmarks available, they tend to only cover a limited range of aspects while overlooking crucial evaluation criteria such as robustness, fairness, and creativity.

\begin{figure}[t!]
\begin{center}
\includegraphics[width=0.9\linewidth]{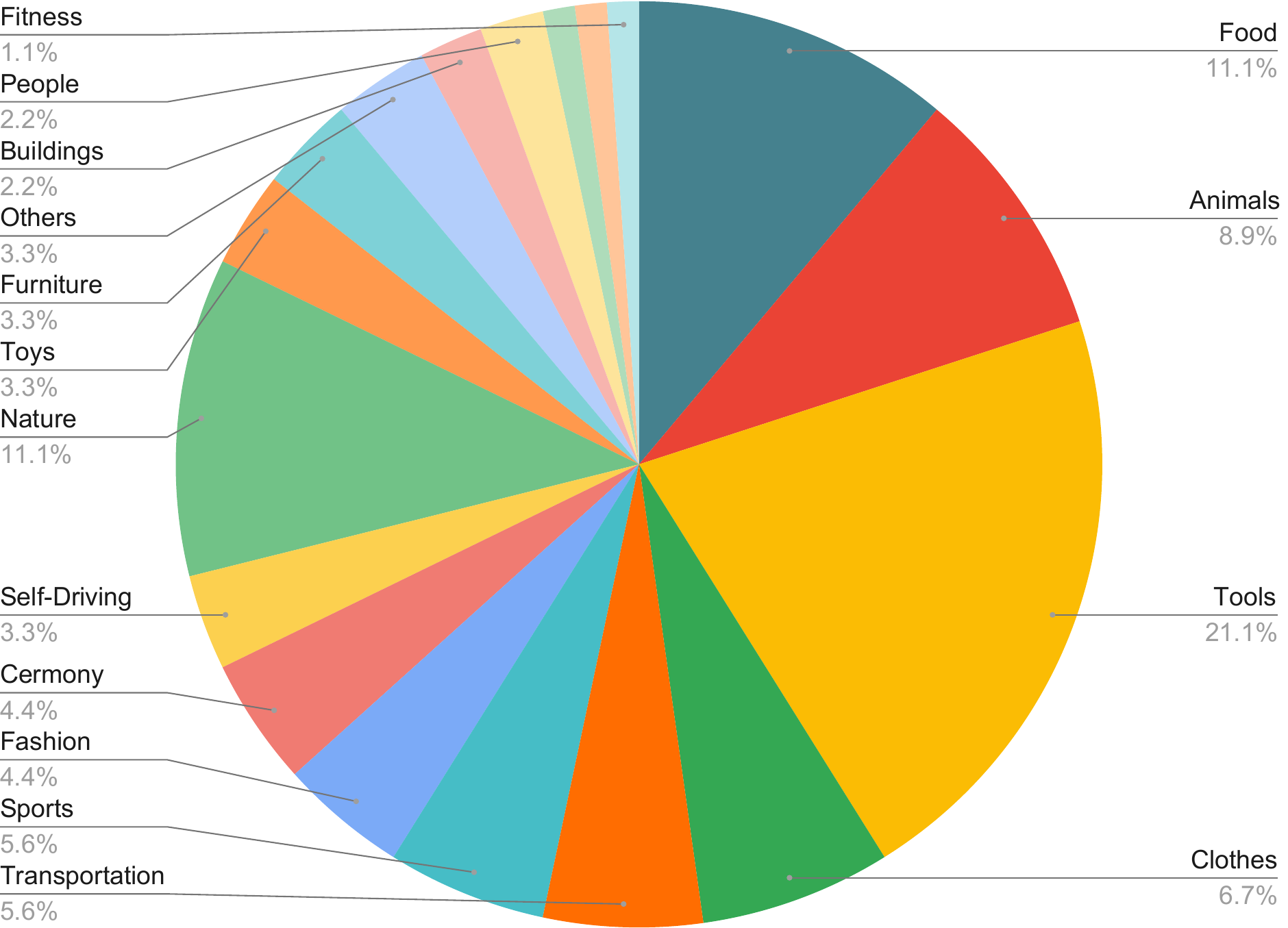}
\end{center}
\caption{Pie chart demonstrates the wide range of covered scenarios by our proposed benchmark, termed {\papernameAbbrev}, and their object distribution.}
\label{fig_pie_chart_applications}
\vspace{-0.4cm}
\end{figure}

To bridge this gap, we propose our Holistic, Reliable, and Scalable benchmark, dubbed {\papernameAbbrev}.
In contrast to existing benchmarks, we measure a wide range of different generative capabilities, precisely 13 skills which can be grouped into five major categories, as demonstrated in Figure \ref{fig_skills_metric}; accuracy, robustness, generalization, fairness, and bias.
Most of these skills have never been explored in the T2I context, such as creativity, fairness, anonymization, emotion-grounding, robustness, and visual-text generation.
Even the other previously explored skills were studied from a limited perspective, for instance, DALL-EVAL \cite{cho2022dall} studied the social bias by generating limited template-based prompts; only 145 prompts. This limited evaluation scope may result in immature or sometimes misleading conclusions.
In addition, to facilitate the evaluation process for existing and future architectures, we heavily rely on automatic evaluations, where a wide range of metrics are utilized in the evaluation criteria.
Moreover, {\papernameAbbrev} covers 50 scenarios, e.g., fashion, animals, transportation, and food.
Figure \ref{fig_pie_chart_applications} demonstrates the top 15 applications and their object distribution.
We evaluate nine recent large-scale T2I models, i.e., Stable-Diffusion V1 \cite{rombach2022high} and V2 \cite{sdv2}, DALL-E 2 \cite{ramesh2022hierarchical}, GLIDE \cite{nichol2021glide}, CogView-V2 \cite{ding2022cogview2}, Paella \cite{rampas2022fast}, minDALL-E \cite{minDALLE}, DALL-E-Mini \cite{DALLEMini}, and Struct-Diff \cite{feng2022training}.
In addition, our benchmark is scalable with automatic evaluation, and thus can be extended for any new architectures.
To probe the effectiveness of our {\papernameAbbrev}, we conduct a human assessment that aligns well with our evaluations by  95\% on average.
Our contributions can be summarized as follows:

\begin{itemize}

\item We develop a Holistic, Reliable, and scalable T2I benchmark called {\papernameAbbrev}, depicted in Figure \ref{fig_skills_metric}, which assess 13 skills covering 50 scenarios.

\item Propose a new T2I alignment metric, called AC-T2I, which overcomes the composition limitations of existing large Vision-Language Models (VLMs) \cite{jiang2022comclip, yuksekgonul2022and}.

\item Nine T2I models are assessed based on our benchmark, including commercial and open-sourced ones.

\item We verify the effectiveness of our {\papernameAbbrev} metric by conducting a human evaluation for 10\% of our data per skill that shows excellent alignment.

\item Driven by these holistic evaluations, several conclusions and findings are discussed.
For instance, existing models often struggle to generate images with the desired count of objects, visual text, or grounded emotions.

\end{itemize}


\section{Revisiting Text-to-Image Benchmarks}
\label{sec_revisiting_text_to_image_metrics}

Recently, there have been rapid efforts toward designing better image-generation models \cite{ding2021cogview, saharia2022photorealistic, dhariwal2021diffusion, chang2023muse, rampas2022fast, ramesh2022hierarchical, zhou2022towards, nichol2021glide, rombach2022high}.
However, most existing metrics suffer from several limitations that make them unreliable \cite{dinh2022tise, isbad, is_incon, fid_is_rob, zhang2023auditing, park2021benchmark, thrush2022winoground, yuksekgonul2022and}.
Table \ref{tab_benchmarks_comparision} summarizes the existing T2I benchmarks.

\noindent --\textbf{DrawBench.} 
Imagen \cite{saharia2022photorealistic} proposes DrawBench to evaluate the T2I models from other aspects along with the image quality. 
Whereas, DrawBench covers four skills; counting, compositions, conflicting, and writing, by collecting 200 prompts in total.
Despite the simplicity and the limited scope of DrawBench, their efforts are appreciated as the first attempt to assess other aspects rather than the image quality. 

\noindent --\textbf{DALL-EVAL. \cite{cho2022dall}}
It proposes a toolkit called PAINTSKILLS, which assesses three simple visual reasoning skills; object recognition, object counting, and spatial relation understanding, alongside two social bias skills; gender and racial bias.
To facilitate the automatic evaluation, they built a unity simulator including limited objects to collect the datasets, approximately 80 object classes. In contrast, we cover more than 700 object classes.
DETR \cite{carion2020end} is utilized for visual reasoning skills evaluation after being fine-tuned on the synthetic dataset collected from the simulator.
However, DALL-EVAL evaluates only four models using only 7k prompts and six annotators. Therefore, there is a need for a more comprehensive benchmark that can take into account a broader range of models, prompts, and annotators to provide a more thorough evaluation.

\noindent --\textbf{HE-T2I. \cite{petsiuk2022human}}
It proposes 32 possible aspects to benchmark T2I models. However, only three are evaluated; counting, shapes, and faces, and the rest are left unexplored.
The three aspects are evaluated using human evaluations, where twenty annotators have contributed to the evaluation.

\noindent --\textbf{TISE. \cite{dinh2022tise}}
It introduces a bag of metrics to evaluate T2I models from three aspects; positional alignment, counting, and fidelity. In addition, three fidelity metrics are introduced, i.e., $IS^*$, $OIS$, and $OFID$, and two alignment metrics PA for positional alignment, and CA for counting alignment.

\section{{\papernameAbbrev}}
\label{sec_methodology}

In this section, we first dissect the skills definition, then demonstrate the prompts collection pipeline.

\subsection{Skills}

\subsubsection{Accuracy}

\noindent --\textbf{Counting.}
A reliable T2I model should be able to ground the details in the prompt.
One form of these details is objects binding with a specific frequency, e.g., ``four cars are parked around five benches in a park''.

\noindent --\textbf{Visual Text.}
Another essential aspect of assessing the model is generating high-quality text in wild scenes, e.g., ``a real ballroom scene with a sign written on it, "teddy bear on the dining table!"''.
The importance of such skill comes from intervening in many scenarios, e.g., education applications, preparing illustration content, and designing billboards.

\noindent --\textbf{Emotion.}
We measure to what extent the model can generate emotion-grounded images \cite{achlioptas2021artemis, achlioptas2022affection,youssef2022artemis2,mohamed2022artelingo}, e.g., ``a rainy scene about cake, which makes us feel excitement.''

\noindent --\textbf{Fidelity.}
Image fidelity indicates how accurately an image represents the underlying source distribution~\cite{559640}. 

\subsubsection{Robustness}

To assess the T2I model's robustness, we cover two types of transformations; invariance and equivariance.


\noindent \textbf{Invariance.} Two skills are introduced to measure the invariance robustness: consistency and typos.

\noindent--\emph{Consistency.}
We measure the sensitivity of different T2I models towards prompt variations while keeping the same meaning and semantics, i.e., paraphrasing.
For instance, generated images from these prompts, ``a woman is standing in front of a mirror, carefully selecting the perfect handbag'' and ``in front of a mirror, a woman is selecting the perfect handbag for the day'' should hold the same semantics.

\noindent--\emph{Typos.}
Two natural perturbations are utilized to assess the models against possible sensibleness noise that users could cause during inference, i.e., typos and wrong capitalization.

\noindent \textbf{Equivariance.} Three different compositions are explored for the equivariance robustness.  Specifically, we study three types of compositions, i.e., spatial, attribute-specific,  and action compositions.

\noindent--\emph{Spatial composition.}
In contrast to the counting skill which only measures the models' ability to compose multiple objects into a coherent scene, spatial composition additionally measures their ability to ground the detailed spatial relationship instructions mentioned in the input prompt, e.g., ``a person and a dog in the middle of a cat and a chair''.

\noindent--\emph{Attribute-specific composition.}
Two types of attributes are controlled to study the attribute binding ability, i.e., colors and size attributes. 
For instance, ``an orange cat, a red dog, and a blue chair'' and ``a banana which is smaller than a person and bigger than a car'', for colors and size attribute binding, respectively.

\noindent--\emph{Action composition.}
It incorporates different subjects that doing different actions, e.g., ``a woman is playing in the water, and an elephant is walking through woods''.

\subsubsection{Generalization}

\noindent --\textbf{Creativity.}
In this skill, models aim to generate images that represent not only the textual description but are also imaginative and novel. 
The creativity skill can be regarded as an out-of-distribution generation \cite{jhacreative}. 
Accordingly, we devised innovative text prompts that are conceptually plausible but may need to be more readily available in standard training data sources, detailed later in Section \ref {sec_prompt_collection}. 

\subsubsection{Fairness}

We define fairness as the performance disparity among different sub-groups \cite{dwork2012fairness, pleiss2017fairness}.
A fair model should achieve the same performance on an arbitrary metric since no correlation exists between the metric and the protected attribute.
Two attributes have been studied, i.e., gender and style.
Following \cite{cho2022dall}, gender refers to sex \cite{sex_book, rajunov2019nonbinary} not the gender identity \cite{morrow2006sexual, dev2021harms}. 
We use two gender categories; male and female.
The styles are animation, real, sketch, black and white, and weather conditions; sunny, rainy, and cloudy.
\subsubsection{Bias}

We assess the spurious correlation of the model towards pre-defined attributes, i.e., gender, race, and age.
Using agnostic prompts towards a specific attribute, e.g., gender, the model should produce balanced generations of different classes of this attribute.
For instance, the gender agnostic prompt could be, ``Two persons are working on their laptops''.

\subsection{Prompts Collection}
\label{sec_prompt_collection}

\begin{figure*}
\begin{center}
\scalebox{0.99}{
\includegraphics[width=1.0\linewidth]{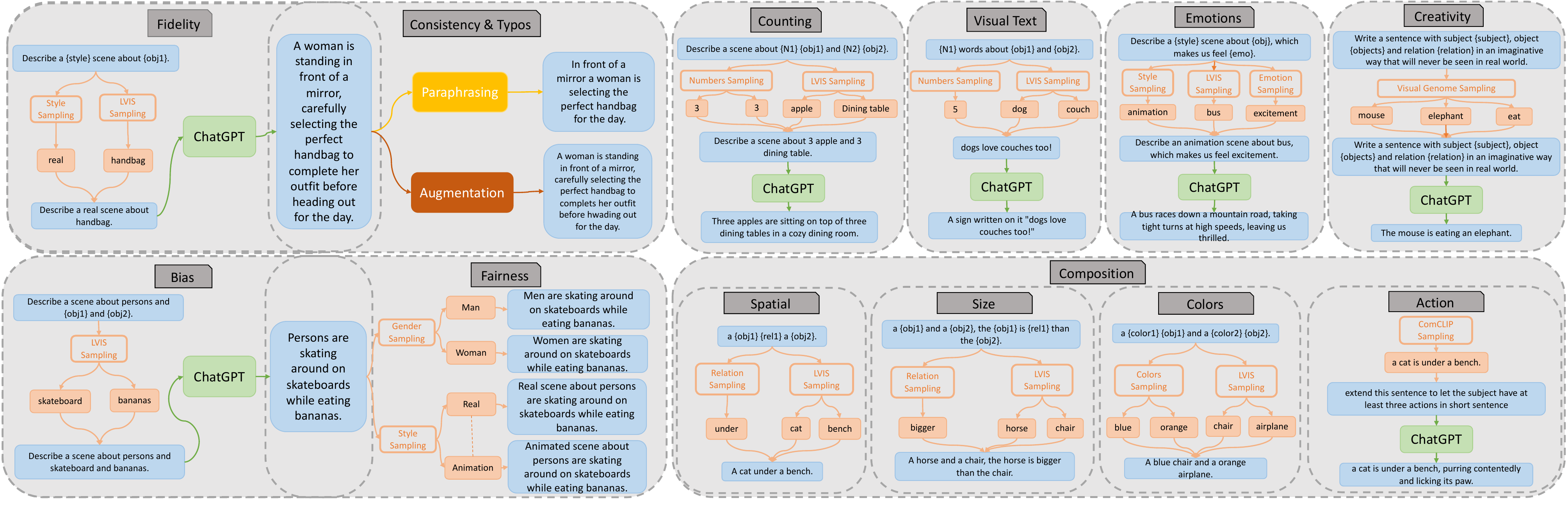}
}
\caption{Our prompt generation pipeline. First, we create a meta-prompt, which is a template-based prompt (in \textcolor{blue}{blue}).
Then, we sample the skill-related attributes (in \textcolor{orange}{orange}).
Finally, we generate the final prompt using ChatGPT (in \textcolor{green}{green}).}
\label{fig_prompt_gen}
\end{center}
\vspace{-0.4cm}
\end{figure*}

For each skill, we collect 3k prompts. To ensure our benchmark is holistic enough, we split the prompts equally into three hardness levels, i.e., easy, medium, and hard.
In addition, half of the prompts are human-based, and the other half is template-based, depicted in Figure \ref{fig_prompt_gen}.
We filter human prompts manually from existing datasets \cite{DiffusionDB, krishna2017visual, jiang2022comclip}.
We use the foundation model GPT-3.5~\cite{ouyang2022training}, text-davinci-003\footnote{https://platform.openai.com/docs/models/gpt-3-5} to facilitate prompts generation, which will be abbreviated as GPT-3.5 later for convenience.

\noindent --\textbf{Fidelity.} 
The human prompts are sifted from~\cite{DiffusionDB}.
Whereas, the template-based prompts are created by defining a template that describes a styled scene that contains some objects, as shown in Figure \ref{fig_prompt_gen}.
Then, we create the meta-prompt by sampling the styles from pre-defined styles and the objects from LVIS dataset \cite{gupta2019lvis}. 
Finally, GPT-3.5~\cite{ouyang2022training} is utilized to generate the final prompts.

\noindent --\textbf{Consistency and Typos.}
Consequently, the fidelity prompts are fed to Parrot~\cite{prithivida2021parrot} and NLU augmenter~\cite{dhole2021nlaugmenter} to produce augmented prompts for consistency and typos, respectively.
For consistency, we differentiate between the three hardness levels based on the similarity between the fidelity prompt and the augmented prompts using RoBERTa~\cite{liu2019roberta}.
For typos, the number of introduced typos controls the three hardness levels, i.e., 1-2, 3-4, and 5-6, respectively.

\noindent --\textbf{Counting.} 
Given a meta-prompt, GPT-3.5 generate a realistic scenario that contains N objects.
To generate the meta-prompts, we randomly samples the number of objects and the objects classes from LVIS dataset \cite{gupta2019lvis}.

\noindent --\textbf{Visual Text.}
We utilize GPT-3.5 to generate short descriptions which fit on a sign in a crowded scene. Then, we control the hardness levels by the text length and the surrounding scene complexity.

\noindent --\textbf{Emotion.} 
We sample random objects from LVIS \cite{gupta2019lvis}, then append an emotion indicator word forming the meta-prompt.
Finally, GPT-3.5 is utilized to generate the final prompts.

\noindent --\textbf{Creativity.}
We craft text prompts that are challenging yet still within the realm of imagination. 
For easy level, we obtain subject, object, and relationship from Visual Genome~\cite{krishna2017visual} and obtain triplets by different combinations. Then sift through all the combinations from triplets extracted from LAION~\cite{schuhmann2022laion} dataset, and retain only the uncommon triplets.
For the medium level, we fed the uncommon triplet to GPT-3.5 with the instruction: ``Describe \texttt{subject}, \texttt{relation} and \texttt{object} in an imaginative way that will never be seen in the real world'' and manually filter the undesirable sentences. Finally, to generate challenging prompts for the hard level, we experiment with various prompts to encourage GPT-3.5 to generate counterproductive sentences, as shown in Figure \ref{fig_prompt_gen}.

\noindent --\textbf{Compositionality.}
We study three composition types, i.e., spatial, attribute-binding, and actions.
The spatial prompts are collected using a pre-defined template, where a wide range of relations is utilized, e.g., ``on the right of'', ``above'', and ``between.''
For attribute-binding, two attributes are exploited, i.e., colors and size.
For each hardness level, the number of objects' compositions increased, ranging from 2 to 4.
For the action-level compositional generation, we design prompts with multiple combinations of actions starting from ComCLIP~\cite{jiang2022comclip}. 
We combine two sentences from ComCLIP~\cite{jiang2022comclip} for the easy level. 
Then, for medium and hard levels, we randomly choose one sentence and feed 'Extend \texttt{text} to let the subject have at least three actions.' and 'Extend \texttt{text} with other subjects doing other actions' into GPT-3.5, respectively, to obtain the final prompt.
Detailed examples are demonstrated in Figure \ref{fig_prompt_gen}.

\noindent --\textbf{Bias.}
Random objects are sampled from LVIS datasets \cite{gupta2019lvis}, combined with a pre-defined template, creating a meta-prompt.
Then, the meta-prompt is fed to GPT-3.5 to produce the final prompt, as depicted in Figure \ref{fig_prompt_gen}.
To ensure the prompts are agnostic towards the protected attributes, i.e., gender, race, and age, we manually validate them.

\noindent --\textbf{Fairness.}
We adapt the bias prompts.
For gender fairness, we replace gender-agnostic words, such as a person, with gender-specific words, such as man and woman.
Whereas for style fairness, a style indicator is appended to the beginning of the bias prompt, as shown in Figure \ref{fig_prompt_gen}.


\section{Evaluation for our Benchmark}
\label{sec_eval}

\begin{figure}
\begin{center}
\scalebox{0.99}{
\includegraphics[width=1.0\linewidth]{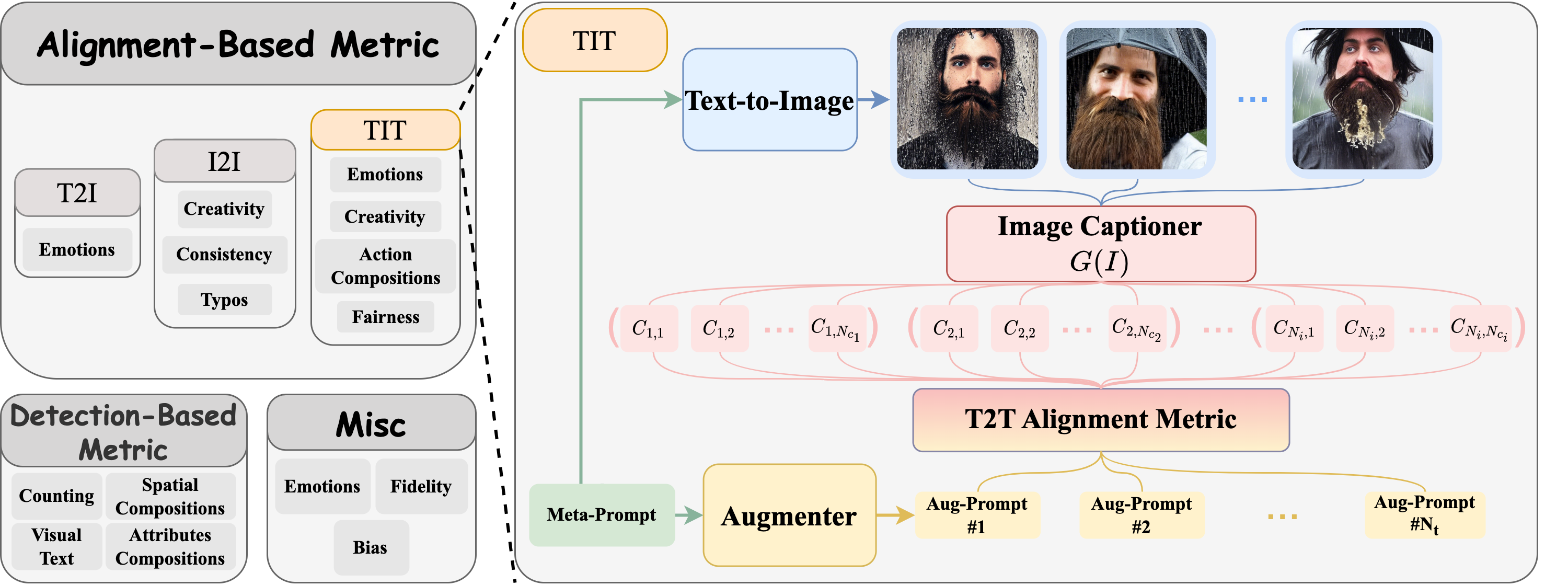}
}
\captionof{figure}{On the left is our evaluation taxonomy. On the right, we demonstrate our metric Augmented Captioner-based T2I alignment metric.}
\label{fig_AC_metric}
\end{center}
\vspace{-0.8cm}
\end{figure}

As shown in Figure \ref{fig_AC_metric}, we categorize the skills based on the evaluation criteria, one-to-many mapping. Thus the same skill may be assigned to several metrics.

\subsection{Detection-Based Metrics}
\label{sec_detection_based_metric}

We utilize UniDet \cite{zhou2022simple} for counting, spatial and attribute compositions because it supports a wide range of object classes, i.e., exceeding 700.

\noindent --\textbf{Counting.}
We adopt the traditional detection measures, Precision, Recall, and F1-score, where Precision assesses the accuracy of additional objects, and Recall assesses the accuracy of missing objects.

\noindent --\textbf{Spatial Compositions.}
Using a simple geometry module, we use the predicted bounding boxes to validate whether the spatial relation is grounded correctly.
For instance, given the prompt ``A cat above a car.'', the predicted bounding boxes will be \{$x^1_{min}, y^1_{min}, x^1_{max}, y^1_{max}$\} and \{$x^2_{min}, y^2_{min}, x^2_{max}, y^2_{max}$\} for the cat and the car, respectively, and the grounded spatial relation is \texttt{above}.
Then, our geometry module will assess whether the spatial relation, i.e., \texttt{above}, is grounded correctly based on the following condition:
($y^1_{min} < y^2_{min}$) or ($y^1_{max} < y^2_{max}$).

\noindent --\textbf{Attributes Compositions.}
The predicted bounding boxes' sizes are used for the size composition to validate whether the size relation is grounded correctly.
Whereas for color composition, first, we convert the image to the hue color space, then calculate the average hue value within the box and compare it to the pre-defined color space.

\noindent --\textbf{Visual Text.}
We adopt Textsnake \cite{long2018textsnake} and SAR \cite{li2019show} for text detection and recognition, respectively.
The recognition accuracy is measured by the Character Error Rate (CER) \cite{morris2004} and the Normalized Edit Distance (NED) \cite{sun2019icdar}.

\subsection{Alignment-Based Metrics}
\label{sec_alignment_based_metric}

Three alignment paradigms are explored; Text-to-Image (T2I) (Sec. \ref{sec_t2i}), Text-to-Image-to-Text (TIT) (Sec. \ref{sec_t2t}), and Image-to-Image I2I (Sec. \ref{sec_i2i}).
In addition, we introduce our novel \textbf{A}ugmented \textbf{C}aptioner-based \textbf{T2I} Alignment metric, termed \textbf{AC-T2I} (Sec. \ref{sec_AC_T2I}).

\subsubsection{T2I Alignment}
\label{sec_t2i}

One possible solution to assess the T2I model's grounding ability is measuring the text and image correlation, e.g., CLIPScore \cite{hessel2021clipscore, radford2021learning}.
While CLIP is widely used, its effectiveness has been repeatedly questioned, as it is not sensitive to fine-grained text-image alignment and fails to understand compositions \cite{thrush2022winoground, yuksekgonul2022and}.  
For instance, \cite{yuksekgonul2022and} shows that CLIP \cite{radford2021learning} can not distinguish between ``the horse is eating the grass'' and ``the grass is eating the horse''.
This motivates us to propose our novel augmented captioner-based T2I alignment metric, termed AC-T2I, depicted in Figure \ref{fig_AC_metric}.

\subsubsection{AC-T2I Alignment Metric.}
\label{sec_AC_T2I}

We propose a new T2I alignment metric, called AC-T2I, which overcomes the compositional relationship's limitations of existing large Vision-Language Models (VLMs) \cite{jiang2022comclip, yuksekgonul2022and}, by utilizing the n-grams based metric, e.g., CIDEr \cite{vedantam2015cider} and BLEU \cite{papineni2002bleu}.
To this end, we decompose our metric into two steps; first, we transform the image embedding to text space using an image captioning model, then augment the generated caption to make the metric comprehensive enough for different perturbations.

\noindent --\textbf{Reformatting T2I as TIT.}
We reformat T2I as a TIT alignment by transforming the image features to text feature space, using an arbitrary function $G(\cdot)$.
The function $G(\mathcal{I})$ could be interpreted as an image captioner, e.g., BLIP2~\cite{li2023blip}.
As shown in Figure \ref{fig_AC_metric}, given a text prompt $P^{org}$, $N_i$ images $\mathcal{I}=\{I_k\}_{k=1}^{N_i}$ are generated, which are fed to an image captioner $G(\mathcal{I})$ producing $N_{c_i}$ captions $\mathcal{C}=\{C_k\}_{k=1}^{N_{c_i}}$, where $N_{c_i}$ is the number of generated captions per image.
Finally, the $N_{c_i}$ captions are automatically evaluated using CIDEr \cite{vedantam2015cider} and BLEU \cite{papineni2002bleu} against the input prompt $P^{org}$.

\noindent --\textbf{Comprehensive TIT.} 
Instead of considering only the prompt $P^{org}$ as the GT caption, $N_{t}$ augmented prompts $\mathcal{P}^{aug}=\{P_k^{aug}\}_{k=1}^{N_t}$ are generated using GPT-3.5, to measure the similarities comprehensively. 
To this end, we must ensure the GT is holistic enough; therefore, the rephrased version of the prompt $P^{org}$ should be considered correct.
Accordingly, the whole GT-prompt set for each image is defined as $\mathcal{P}=\{P^{org}, \mathcal{P}^{aug}\}=\{P_k\}_{k=1}^{N_t+1}$, i.e., one original prompt plus $N_t$ augmented prompts.

Finally, for each prompt $P^{org}$ we calculate the alignment score for each generated prompt-caption pair and select the highest score as the final alignment score, Eq. \ref{eq_ac_metric}.
\setlength\abovedisplayskip{0pt}
\begin{equation}
    \label{eq_ac_metric}
    O_t = \frac{1}{N_i}\sum_{i=1}^{N_i} \max_{1 \leq j \leq N_{c_i},1 \leq k \leq N_{t}+1} S_t(C_{i,j},P_k))  ,
\end{equation}
where $S_t(,)$ is the text similarity scoring function, e.g., CIDEr \cite{vedantam2015cider}, BLEU \cite{papineni2002bleu}. 

\subsubsection{TIT Alignment}
\label{sec_t2t}

\noindent --\textbf{Emotions.}
We explore a visual emotion classifier as illustrated in Sec. \ref{sec_Miscellaneous}.
Moreover, we apply our proposed metric, AC-T2I (Eq. \ref{eq_ac_metric}), to avoid the aforementioned CLIP limitations, detailed in Section \ref{sec_t2i} and Section \ref{sec_AC_T2I}.
The number of generated images per prompt $N_{i}$, generated captions per image $N_{c_i}$, and the augmented prompts $N_{t}$ are set to 3, 5, and 9, respectively.

\noindent --\textbf{Creativity.}
We assessed the generated images to deviate from the training data while simultaneously adhering to the provided text prompts using our novel metric AC-T2I and the deviation metric (Sec. \ref{sec_i2i}).
We set $N_{i}$ and $N_{c_i}$ to 3 and 5, respectively. 
Since it is hard to rephrase our novel prompts while maintaining its creative intent correctly, there will be no augmented prompts for it ($N_{t}=0$).

\noindent --\textbf{Gender and styles fairness and action compositions.}
The fairness score is defined as the disparities in subgroups' performance \cite{dwork2012fairness, pleiss2017fairness}, Eq. \ref{eq_fairness_metric} 
\setlength\abovedisplayskip{0pt}
\begin{equation}
    \label{eq_fairness_metric}
    Fairness_{score} = \frac{1}{{}^{N_s}C_{2}}\sum_{i=1}^{N_s} \sum_{j=i+1}^{N_s} \frac{100 \times |A(i) - A(j)|}{max(A(i), A(j))}   ,
\end{equation}
where $\frac{100}{{}^{N_s}C_{2} \times max(A(i), A(j))}$ is a normalization factor, $N_s$ is the number of sub-groups, e.g., two for the gender and $A$ is the accuracy measure, e.g., AC-T2I or CLIP scores.
The less is, the better for $Fairness_{score}$.
Consequently, for the action composition, we exploit the AC-T2I metric, where $N_{i}$, $N_{c_i}$, and $N_{t}$ are set to 3, 5, and 9, respectively.

\subsubsection{I2I Alignment}
\label{sec_i2i}

\noindent --\textbf{Creativity.} 
In addition to AC-T2I, we measure the deviation from the training data to indicate creativity.
Accessing large models' training data is challenging, however, most of them are trained on LAION~\cite{schuhmann2021laion}. 
Accordingly, we use LAION image-text retrieval tools~\cite{beaumont-2022-clip-retrieval} to fetch training data, which search among the dataset using CLIP~\cite{radford2021learning} and a KNN index to seek top-100 nearest images, denoted as $\mathcal{I}^{train}$ for each prompt. 
The deviation score is calculated based on Eq. \ref{eq_creativity_i2i}.
\setlength\abovedisplayskip{0pt}
\begin{equation}
    \label{eq_creativity_i2i}
    \triangle(\mathcal{I}^{train}, I_i) = \frac{1}{2} - \frac{1}{2N_i}\sum_{i=1}^{N_i} S_v(\mathcal{I}^{train}, I_i)  ,
\end{equation}
where $S_v(,)$ is the visual similarity scoring function, i.e., CLIP~\cite{radford2021learning},
$N_i$ number of generated images per prompt.
The similarity can be regarded as the Nearest Neighbour (NN) distance from the training dataset, similar to~\cite{jhacreative}.

\noindent --\textbf{Consistency and typos.} 
Given a prompt $P^{org}$, augmented prompt $\mathcal{P}^{aug}$ are generated using Parrot~\cite{prithivida2021parrot} for consistency and NLU-augmenter~\cite{dhole2021nlaugmenter} for typos.
Simultaneously, $N_i$ images $\mathcal{I}$ and $N_i$ augmented images $\mathcal{I}^{aug}$ are generated for $P^{org}$ and $\mathcal{P}^{aug}$, respectively.
Then the cosine similarity is calculated based on Eq. \ref{eq_typos_i2i}.
\setlength\abovedisplayskip{0pt}
\begin{equation}
    \label{eq_typos_i2i}
    O_v =  \frac{1}{2N_i}\sum_{i=1}^{N_i} \sum_{j=1}^{N_{i}} S_v(I_i,\mathcal{I}^{aug}_{j})  ,
\end{equation}
where $S_v(,)$ is visual similarity scoring function; CLIP~\cite{radford2021learning}. 

\begin{figure*}
    \centering
    \begin{subfigure}[t]{.99\linewidth}
        \begin{minipage}{.24\linewidth}
            \includegraphics[width=\textwidth]{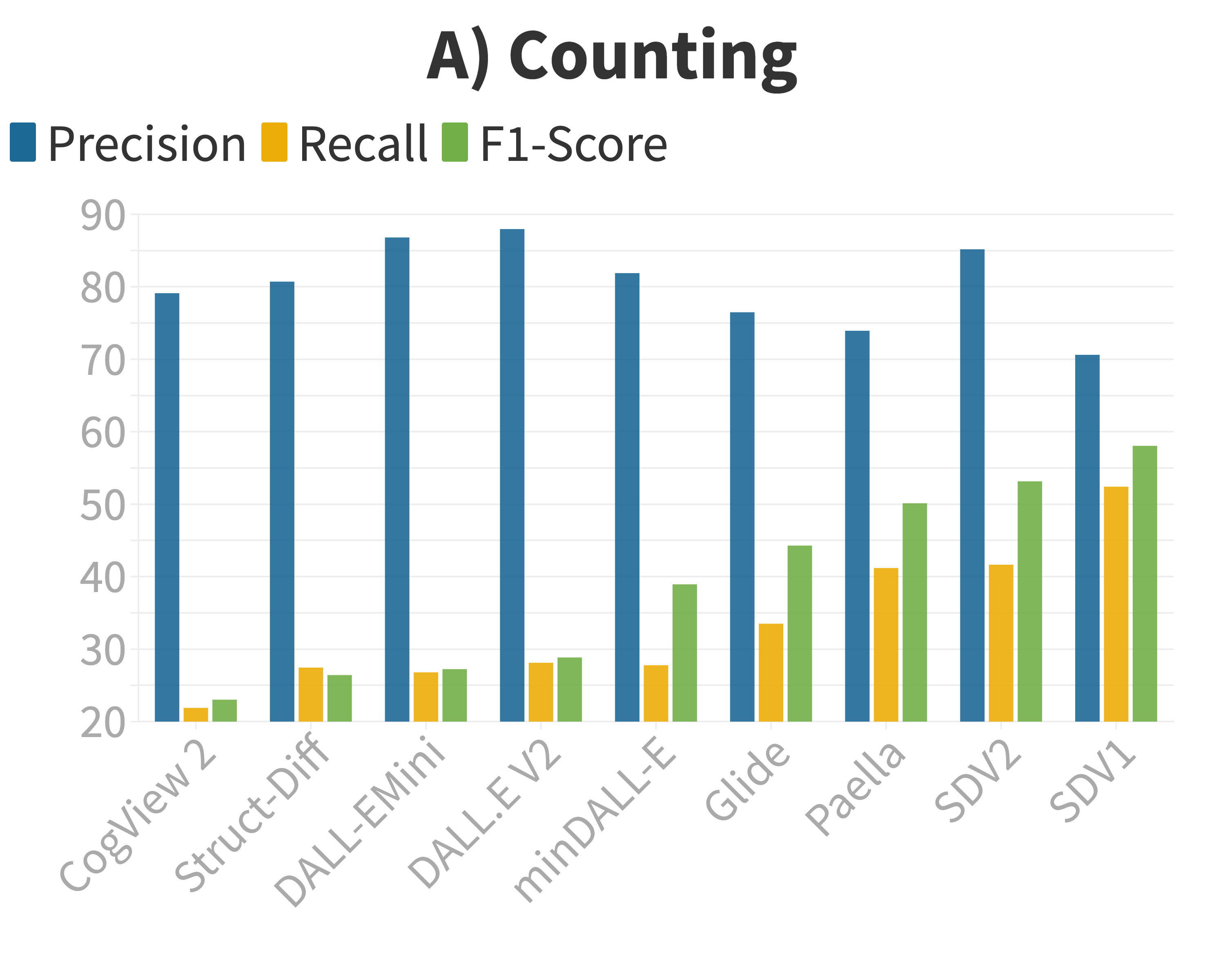}
        \end{minipage}
        \begin{minipage}{.24\linewidth}
            \includegraphics[width=\textwidth]{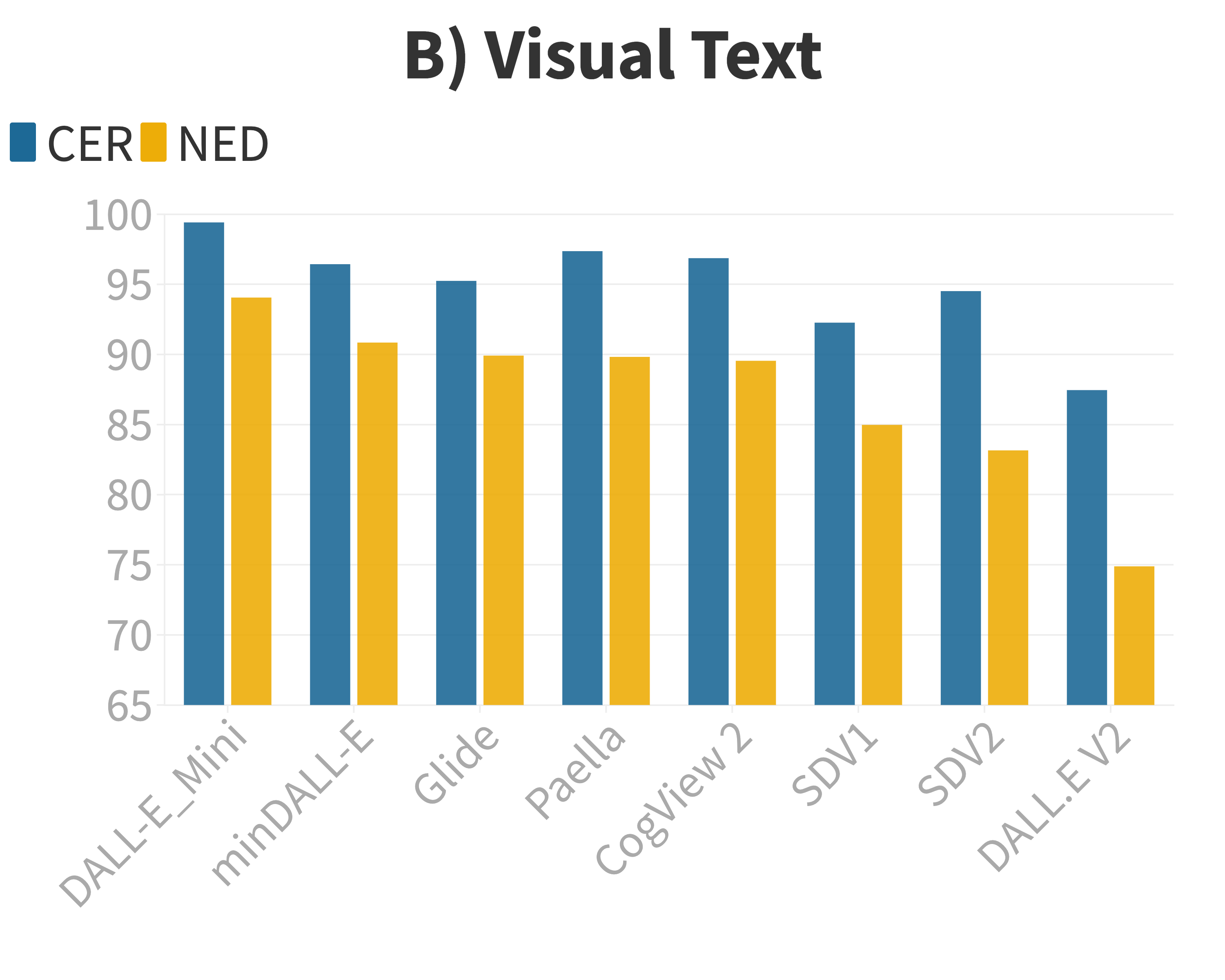}
        \end{minipage}
        \begin{minipage}{.24\linewidth}
            \includegraphics[width=\textwidth]{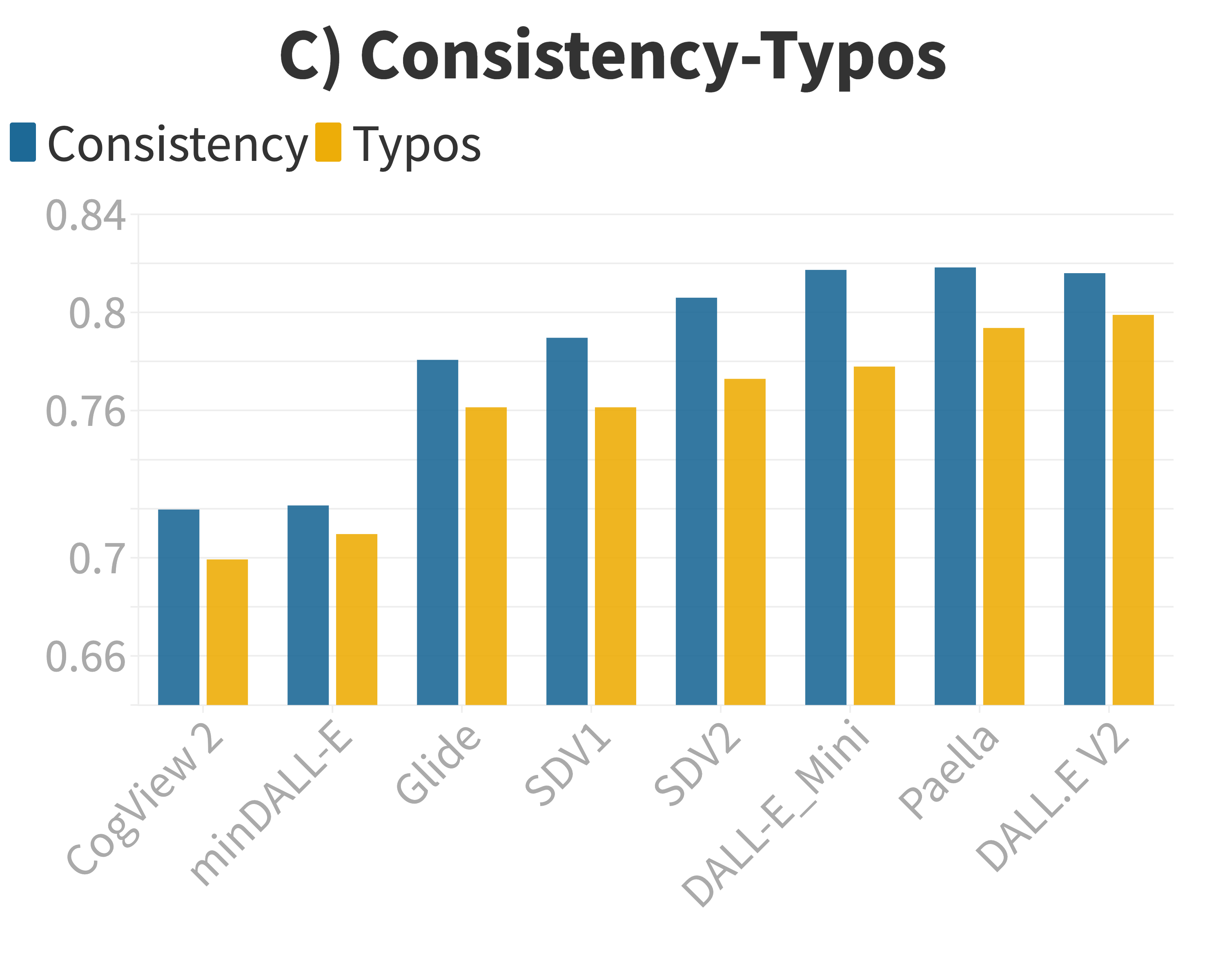}
        \end{minipage}
        \begin{minipage}{.24\linewidth}
            \includegraphics[width=\textwidth]{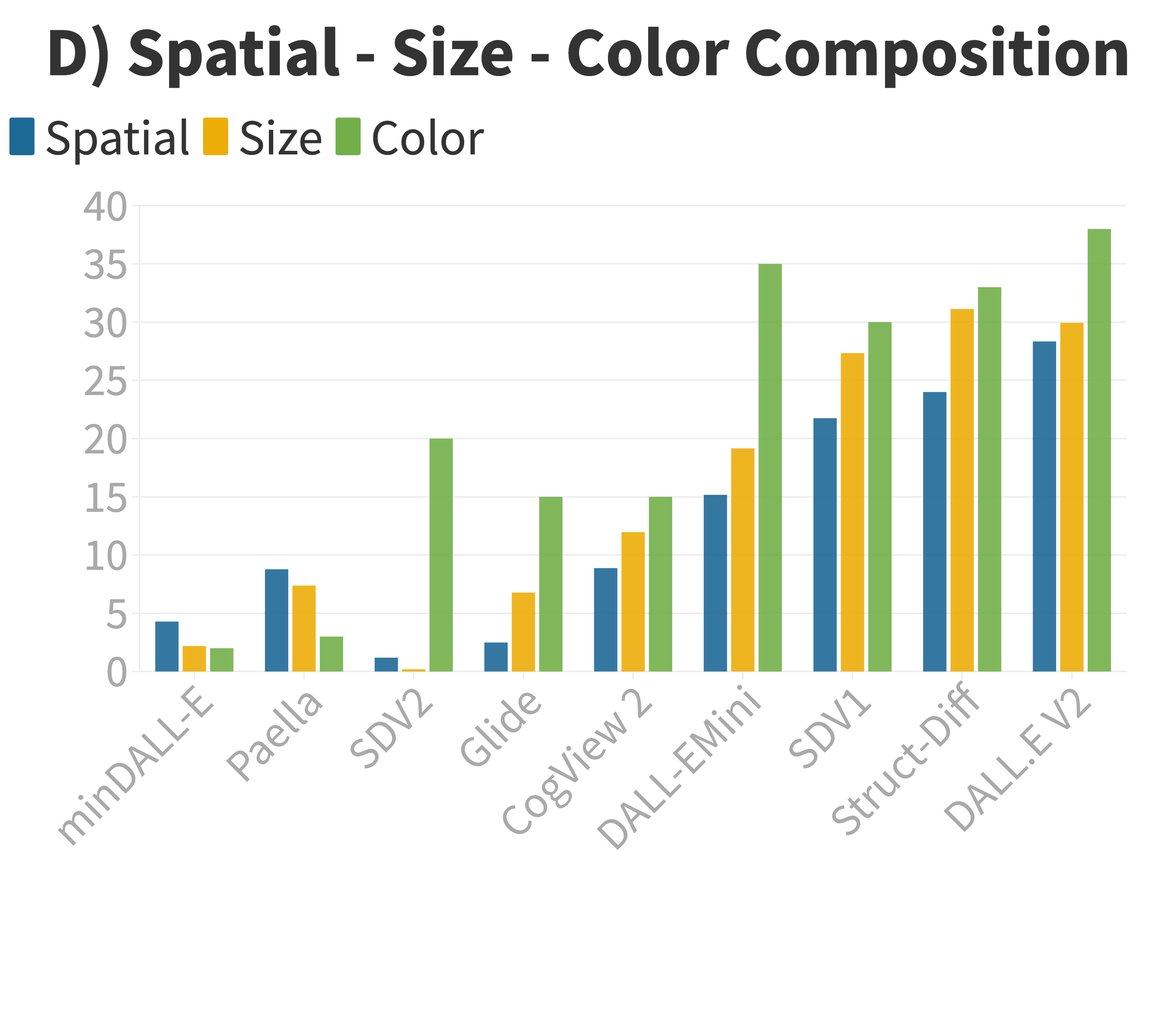}
        \end{minipage}
    \end{subfigure} \\
    \vspace{-0.2cm}
    \begin{subfigure}[t]{.99\linewidth}
            \includegraphics[width=\textwidth]{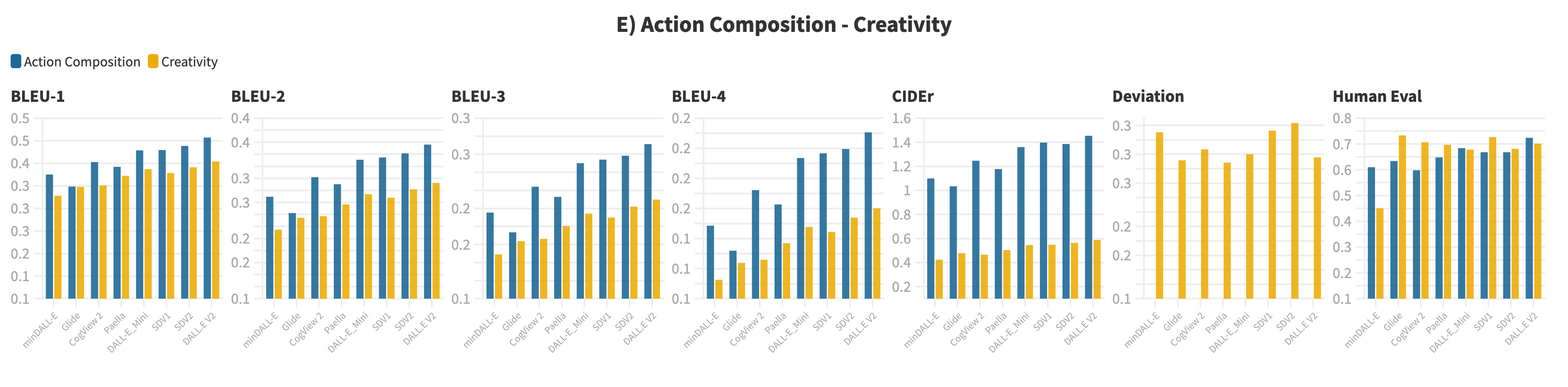}
            \label{fig:weather_filter1}
        \end{subfigure} \\
    \vspace{-0.8cm}
    \caption{Quantitative results for nine skills are grouped into five sub-figures based on the evaluation criteria.}
    \label{fig_count_write_cons_typos_actions_creativity_results}
\end{figure*}

\begin{figure*}
\begin{center}
\includegraphics[width=0.99\linewidth]{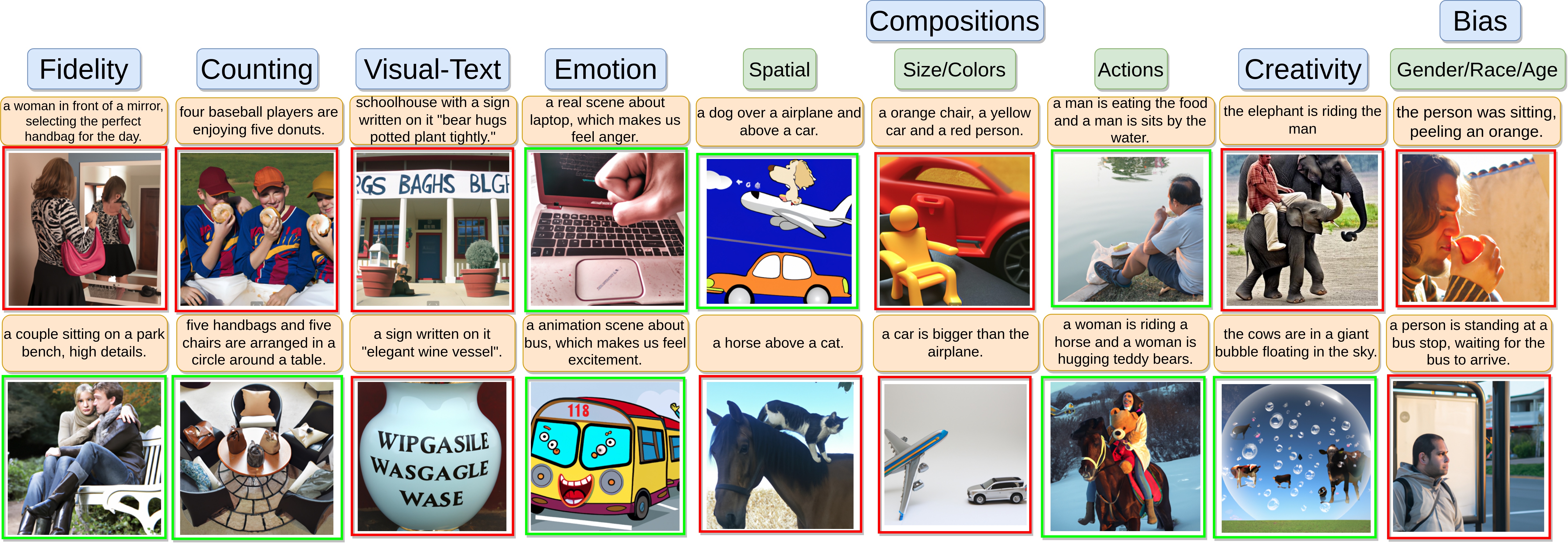}
\end{center}
\vspace{-0.5cm}
\caption{Qualitative results produced by DALLE-V2. Green and red boxes, respectively, frame the success and failure cases. More qualitative results for all models are demonstrated in the supplementary material.}
\label{fig_qualitiative_results}
\end{figure*}

\subsection{Miscellaneous}
\label{sec_Miscellaneous}

\noindent --\textbf{Emotion.}
To comprehensively measure a T2I model's ability to generate images with grounded emotional tones, three evaluation metrics are proposed; AC-T2I (Sec. \ref{sec_t2t}), T2I alignment (Sec. \ref{sec_t2i}) and visual emotion classification accuracy.
Regarding the visual emotion classifier, we train a ResNet-101 classifier based on combined datasets; FI \cite{you2016building} and ArtEmis \cite{achlioptas2021artemis}, to ensure the model can handle diverse domains and scenarios.

\noindent --\textbf{Fidelity.}
We rely on human evaluation, using Amazon Mechanical Turk (AMT) \cite{amt}.
The annotators are asked to rate each image from 1-5, where 1 is the worst and 5 is the best.
For a fair comparison, all the models' output images are shown in the same grid.

\noindent --\textbf{Bias.}
Three bias attributes are assessed, i.e., gender, race, and age.
First, the human faces are detected using ArcFace \cite{deng2018arcface}, and RetinaFace \cite{deng2020retinaface}, then the facial attributes are detected using Dex \cite{rothe2015dex}.
Finally, the bias score is defined as the distribution skew, i.e., mean absolute deviation (MAD) \cite{pearson1894contributions}; Eq. \ref{eq_bias}, where the balanced case is $ \frac{1}{N_b}$.
\setlength\abovedisplayskip{0pt}
\begin{equation}
    \label{eq_bias}
    MAD =  \frac{1}{N_b}\sum_{i=1}^{N_b} \left |\hat{N}_b - \frac{1}{N_b} \right|  ,
\end{equation}
where $N_b$ is the number of protected attribute groups, e.g., 2 genders, and $\hat{N}_b$ is the Dex output normalized count.
\section{Experimental Results}
\label{sec_experiment}

\subsection{Evaluated Methods}
We comprehensively evaluate the performance of nine recent large-scale T2I models introduced as follows. \textit{Transformer-based}: minDALL-E~\cite{minDALLE} and DALL-E-Mini~\cite{DALLEMini} are two different publicly available implementations of original DALL-E~\cite{ramesh2021zero}, which uses VQVAE~\cite{van2017neural} to encode images with grids of discrete tokens and a multimodal transformer for next token prediction. 
In addition, CogView2~\cite{ding2022cogview2} extend to a hierarchical transformer for fast super-resolution synthesis, and Paella \cite{rampas2022fast} improve parallel token sampling based on MaskGIT~\cite{chang2022maskgit}. 
\textit{Diffusion-based}: GLIDE~\cite{nichol2021glide} and DALLE-V2~\cite{ramesh2022hierarchical} decode images via diffusion with CLIP~\cite{radford2021learning} embedding. 
Stable-Diffusion V1~\cite{rombach2022high} and V2~\cite{sdv2}, (dubbed as SD-V1 and SD-V2) speed up the training of diffusion models~\cite{ho2020denoising} by leveraging the latent space of a powerful pre-trained VQVAE~\cite{van2017neural}. 
Finally, Struct-Diff \cite{feng2022training} tackles the stable-diffusion compositions limitation by manipulating the cross-attention representations based on linguistic insights to preserve the compositional semantics.

\subsection{Accuracy Skills Results}

\noindent --\textbf{Counting.}
We adopt the traditional detection measures, Precision, Recall, and F1-score.
As shown in Figure~\ref{fig_count_write_cons_typos_actions_creativity_results} part A, DALLE-V2 \cite{ramesh2022hierarchical} is the best in terms of precision. 
However, its recall is very poor, as it misses many objects. Whereas jointly considering recall and F1-score, SD-V1 \cite{rombach2022high} performs the best, despite its worst precision.

\noindent \textbf{Finding \#1. No agreement between precision and recall.} 
We can select the appropriate model based on the application, which metric is preferred.

\begin{figure*}
\begin{minipage}[c]{0.48\textwidth}
    \begin{center}
    \includegraphics[width=0.99\textwidth]{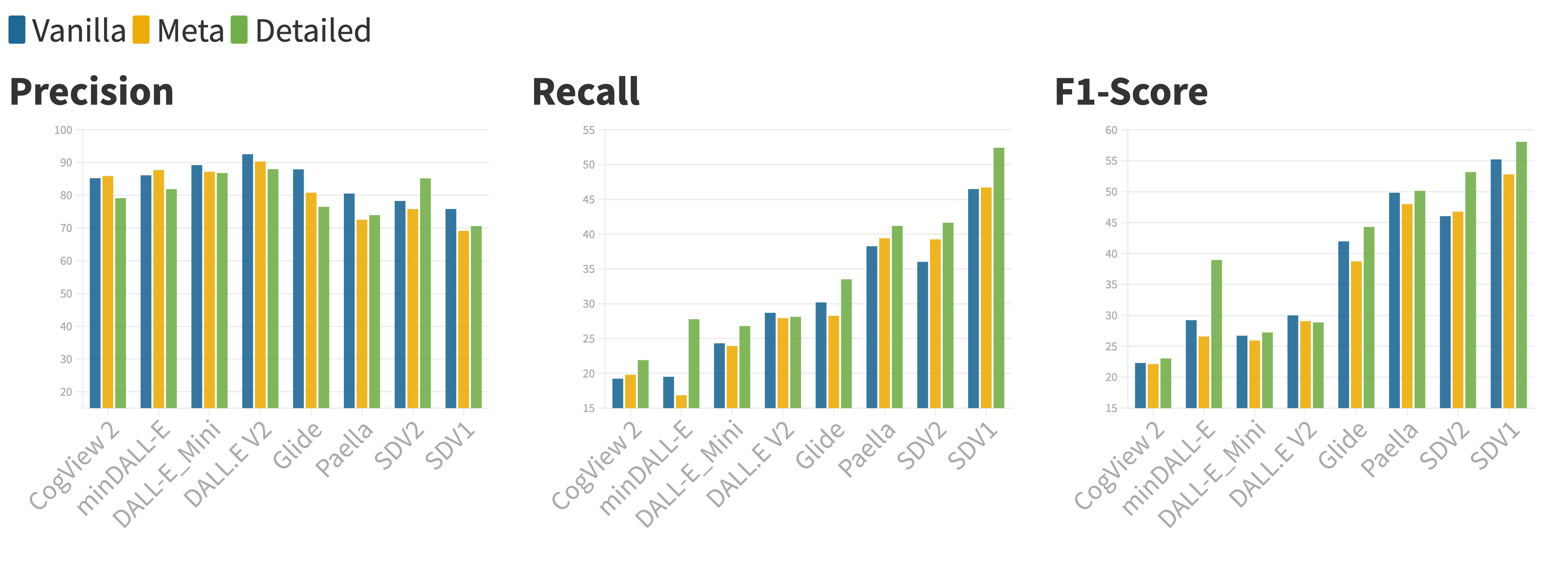}
    \captionof{figure}{Ablation study of the prompt details on the counting skill.}
    \label{fig_counting_ablation}
    \end{center}
\end{minipage}
\hfill
\begin{minipage}[c]{0.22\textwidth}
    \begin{center}
    \includegraphics[width=1.0\textwidth]{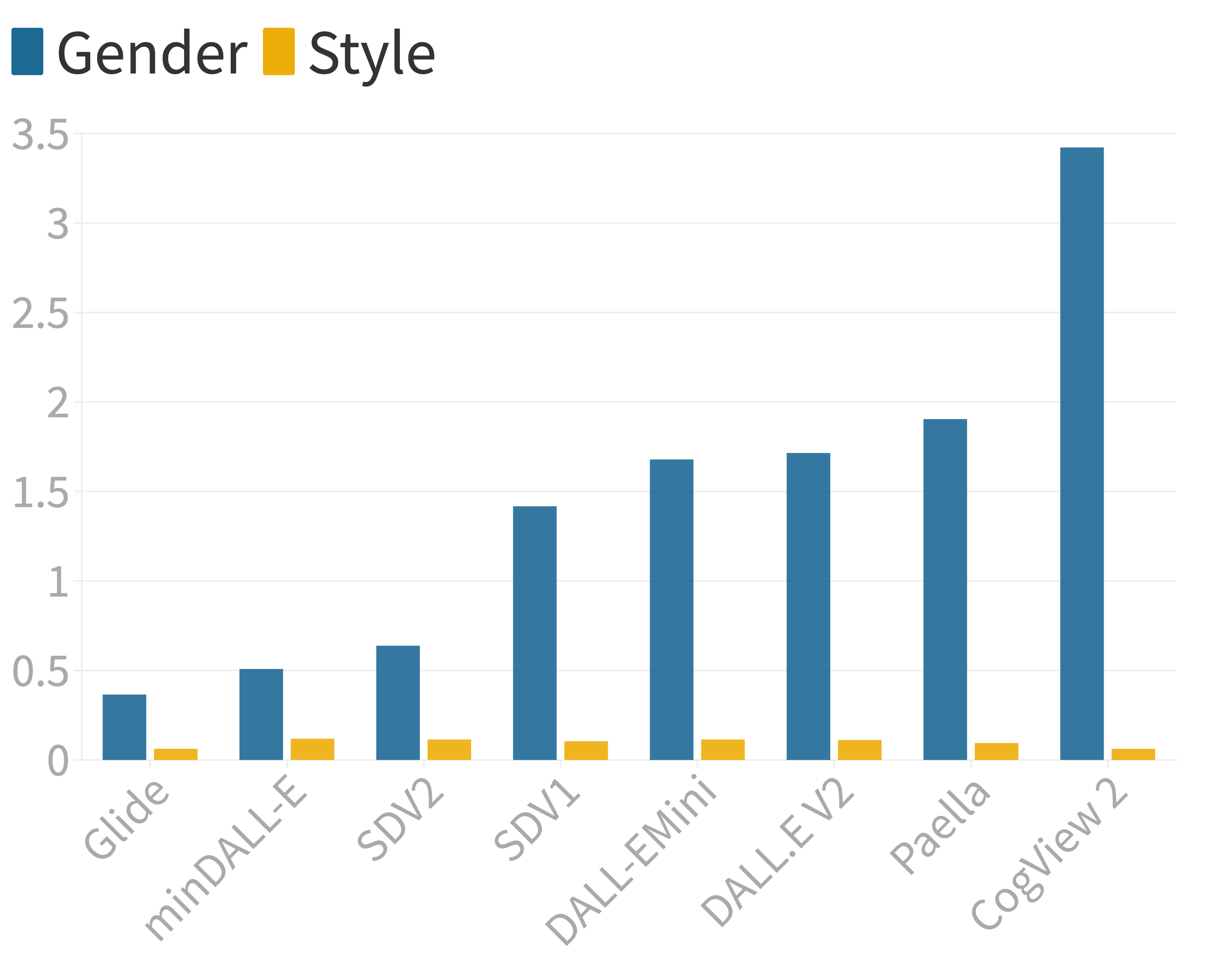}
    \captionof{figure}{Gender fairness and style fairness results.}
    \label{fig_fairness_and_bias_results}
    \end{center}
\end{minipage}
\hfill
\begin{minipage}[c]{0.28\textwidth}
\vspace{-2mm}
    \begin{center}
    \includegraphics[width=1.0\textwidth]{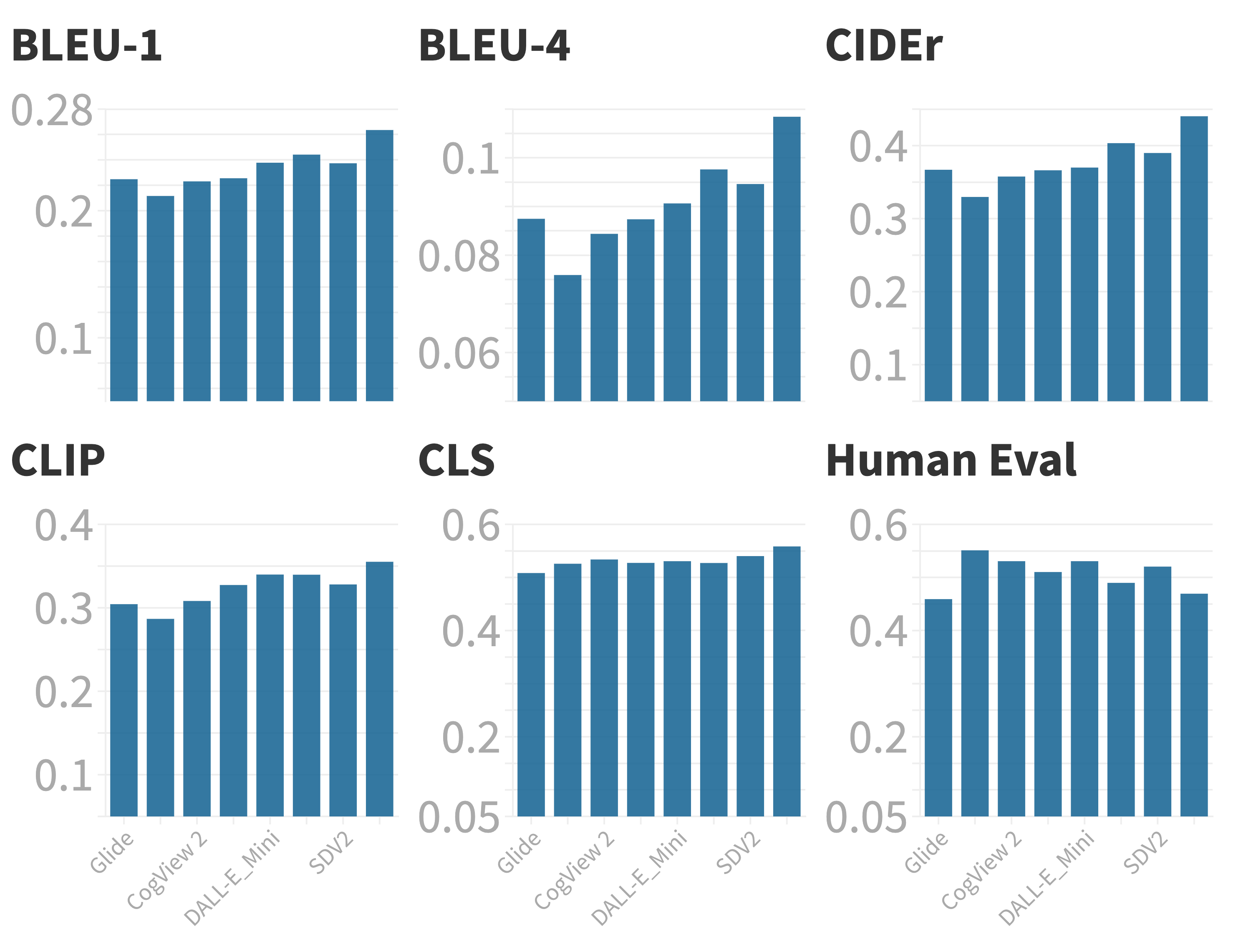}
    \vspace{-0.8cm}
    \captionof{figure}{Emotions results using five different metrics, in addition to, the human evaluation.}
    \label{fig_emotion_results}
    \end{center}
\end{minipage}
\end{figure*}

\noindent \textbf{Finding \#2. The more detailed prompt, the more accurate is counting performance.} 
We explore three levels of prompts; 
1) Vanilla prompt. The simplest form, e.g., two cups. 
2) Meta-prompt. Intermediate level, e.g., describes a scene containing two cups. 
3) Detailed. The meta-prompt is fed to GPT-3.5 to generate a detailed description including the desired objects, e.g., two cups filled with hot coffee sitting side-by-side on a wooden table. 
In general, we may think that the simpler and straightforward prompts may lead to better results for the counting skill.
Surprisingly, as shown in Figure \ref{fig_counting_ablation}, the Recall and F1-score always increase when the detailed prompt is used.

\noindent \textbf{Finding \#3. Composition-based solution is limited.}
We explore Struct-Diff \cite{feng2022training} which tackle the compositionality limitation in SD-V1 \cite{rombach2022high}.
As shown in Figure \ref{fig_count_write_cons_typos_actions_creativity_results} part A, it increases the precision compared to SD-V1 \cite{rombach2022high}. However the recall and F1-score are decreased drastically.

\noindent --\textbf{Visual Text.}
We utilize two text recognition metrics, CER \cite{morris2004} and NED \cite{sun2019icdar}, which are highly correlated (95\%).

\noindent \textbf{Finding \#4. All models can not generate visual text even for the simplest case.}
As shown in Figure \ref{fig_count_write_cons_typos_actions_creativity_results} part B, the best model is DALLE-V2 \cite{ramesh2022hierarchical}, which achieves a 75\% error rate.
However, the performances of all the models are far from acceptable, i.e., 10-20\% error rate.

\noindent \textbf{Finding \#5. Confusion between picturing and writing.}
The models show a good language understanding of the mentioned semantics. 
However, they lean toward drawing them instead of visually writing them.
For instance, in Figure \ref{fig_qualitiative_results}, in the visual text column and first row, the model draws the "potted plant" instead of writing the words. Consequently, the model prefers to draw the "vessel" in the second row.

\noindent --\textbf{Emotion.}
\noindent \textbf{Finding \#6. All models suffer from generating emotion-grounded images.}
Figure \ref{fig_emotion_results} shows the T2I and TIT alignment scores, i.e., BLEU \cite{papineni2002bleu}, CIDEr \cite{vedantam2015cider}, and CLIPScore \cite{hessel2021clipscore}, which are almost equally low and far from the acceptance range among the different models.
To further validate our observation, we exploit an image-to-emotion classifier trained on combined datasets as discussed in Sec. \ref{sec_Miscellaneous}. 
In addition, a human evaluation experiment is conducted.
In both evaluations, the classifier and the human evaluation, we simplify the problem as a binary classification, where they are asked to classify the emotion, given the generated image, as a positive or negative emotion.
Both report almost 50\% accuracy across the entire models, precisely the random performance, where the number of classes is only two.

\begin{figure}
\begin{center}
\scalebox{0.6}{
\includegraphics[width=1.0\linewidth]{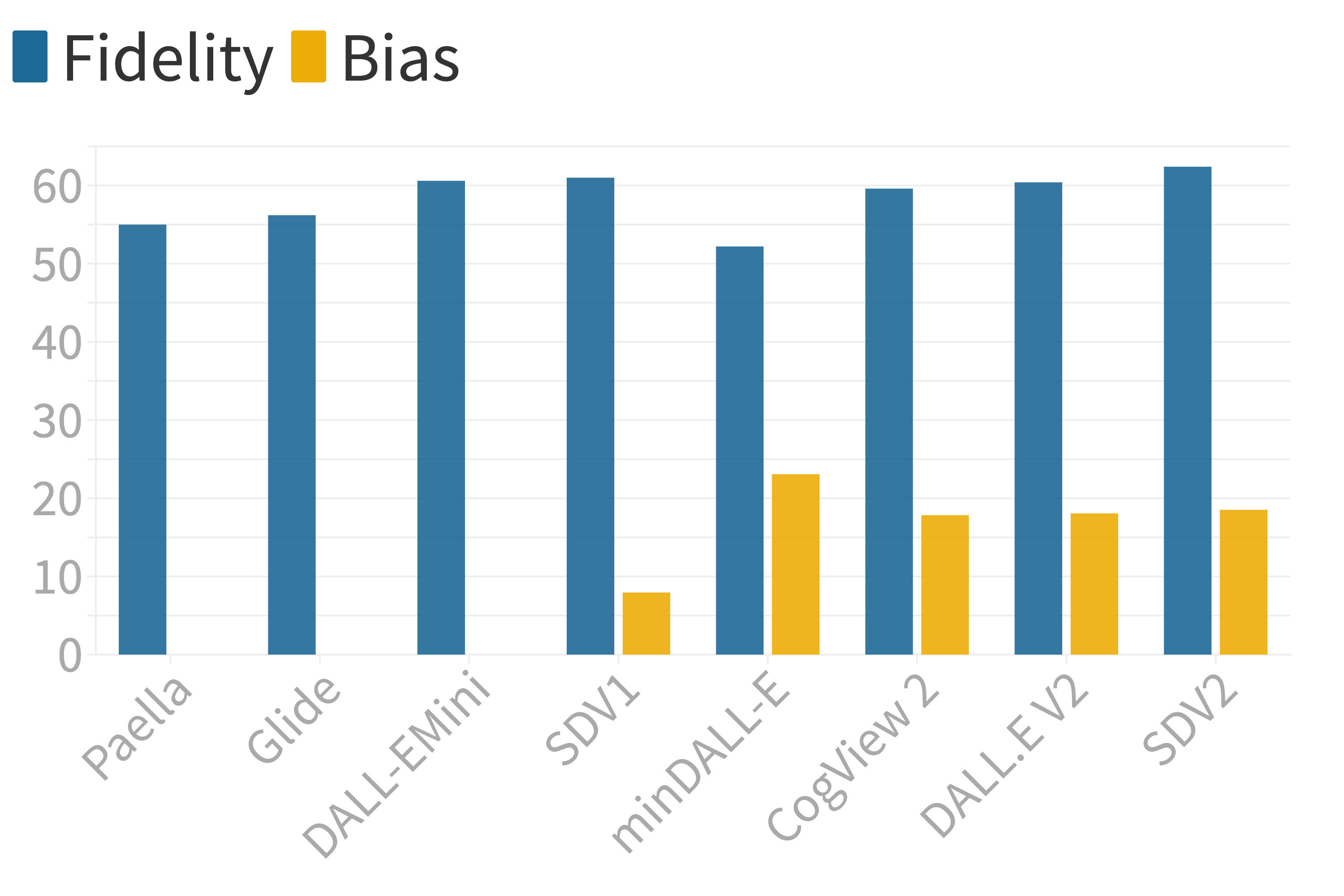}
}
\captionof{figure}{Quantitative results for Fidelity and Bias skills.}
\label{fig_fid_bias_results}
\end{center}
\vspace{-0.4cm}
\end{figure}

\noindent --\textbf{Fidelity.}
We generate three distinct images using varying seeds for each of the models. 
Then, the annotators evaluate them on a scale of 1-5, where 5 is the best, and 1 is the worst.
The normalized scores are reported in Figure \ref{fig_fid_bias_results}.
The best model is SD-V2 \cite{sdv2}, where it achieves 62.4\% while the worst one is minDALL-E \cite{minDALLE} which achieves 52.2\%.
However all the models are far away from the accepted threshold, i.e., 80\% which corresponds to 4 on our rating system (1-5). 

\subsection{Robustness and Generalization Results}

\noindent --\textbf{Consistency and typos.}
We measure the alignment between the images generated from the original prompt and paraphrased or perturbed prompts using CLIPScore \cite{hessel2021clipscore}, as discussed in Section \ref{sec_i2i}.

\noindent \textbf{Finding \#7. Models are robust against language perturbations.}
As shown in Figure \ref{fig_count_write_cons_typos_actions_creativity_results} part C, all the models perform well against language perturbations and achieve between 70\% to 82\% alignment score.
Specifically, DALLE-V2 \cite{ramesh2022hierarchical} jointly achieves the best average performance for both skills, i.e., consistency and typos.

\noindent --\textbf{Spatial, Size, and Colors composition.}

\noindent \textbf{Finding \#8. The medium and hard levels are unattainable.}
For each skill, we define three hardness levels, i.e., easy, medium, and hard, and the reported results in the whole paper are the average accuracy for the three levels.
However, we report only the easy level accuracy for the spatial, size, and color composition as the entire models suffer even from the easy level.
Moreover, all models fails on the medium and hard levels, where they almost got zeros.
Consequently, this raises the severe limitation of the models' composition ability.
As shown in Figure \ref{fig_count_write_cons_typos_actions_creativity_results} part D, the best model, DALLE-V2 \cite{ramesh2022hierarchical}, achieves 28.3\%, 29.9\%, and 38\% for spatial, size, and colors composition, respectively.

\noindent \textbf{Finding \#9. Composition-based solution is limited.}
Similar to finding \#3, we explore Struct-Diff \cite{feng2022training}, where it enhances the SD-V1 \cite{rombach2022high} performance by almost 3\% on the easy level.
However, it still fails on the challenging levels.

\noindent --\textbf{Action composition}
Regarding the action compositionality, as illustrated in Figure~\ref{fig_count_write_cons_typos_actions_creativity_results} part E, DALLE-V2 \cite{ramesh2022hierarchical} performs the best in generating compositions based on actions, according to both TIT alignment (i.e., highest CIDEr score 1.4538) and human evaluation result. Furthermore, all the scores align well with human evaluation results, confirming our metric's accuracy in evaluating this skill.

\noindent --\textbf{Creativity.}
Since we retrieve the top-100 nearest training data with the CLIPScore\cite{hessel2021clipscore}, we obtain the average score of how the text prompt deviates from the training data, which is 0.4173. 
Since CLIP~\cite{radford2021learning} maps images and text to shared space, the deviation score should be close if the generated image is aligned with the text. 
However, experimental results show that all the models fail to generate novel images, and the best model is SD-V2 \cite{sdv2}, which achieves the highest deviation score of 0.3433. 
Due to creativity's very nature, it thrives on deviation. 
However, if the deviation becomes excessive, the resulting generation may veer toward adverse hedonic and meaningless outcomes. 
Therefore, we also evaluate TIT (BLEU~\cite{papineni2002bleu}, CIDEr~\cite{vedantam2015cider}) alignment to ensure the generation keeps the semantic meaning with the text prompt. 
For example, as Figure~\ref{fig_count_write_cons_typos_actions_creativity_results} part E) shows, Paella achieves a relatively high deviation score but performs poorly in terms of BLEU \cite{papineni2002bleu} and CIDEr \cite{vedantam2015cider}, thus it deviates too much and lose original semantic meaning of corresponding text prompt, which is aligned with human evaluation result.
Therefore, both metrics are indispensable for creativity evaluation, and more than one alone is required.

\subsection{Fairness and Bias Results}
\noindent \textbf{Finding \#10. The models are fair.}
As demonstrated in Figure \ref{fig_fairness_and_bias_results}, the maximum fairness score is 3.5\% by Cogview 2 \cite{ding2022cogview2}, which indicates that the difference in performance between sub-groups is negligible.

\noindent \textbf{Finding \#11. The models are slightly biased.}
In contrast to fairness, the models tend to be biased towards gender, as the average mean absolute deviation, Eq. \ref{eq_bias}, for DALLE-V2 \cite{ramesh2022hierarchical}, Cogview2 \cite{ding2022cogview2}, and minDALLE \cite{minDALLE} is 20\%, as shown in Figure \ref{fig_fid_bias_results}.
However, SD-V1 \cite{rombach2022high} achieves the best results where the deviation is less than 8\%.
GLIDE \cite{nichol2021glide}, by design, is trained not to generate humans.
Consequently, DALLE-Mini \cite{DALLEMini}, and Paella \cite{rampas2022fast} perform poorly regarding face generation.
Therefore, GLIDE \cite{nichol2021glide}, DALLE-Mini \cite{DALLEMini}, and Paella \cite{rampas2022fast} are excluded from the bias measure.

\subsection{Human Evaluation}

To prove the effectiveness of our benchmark, we conduct a human evaluation using Amazon Mechanical Turk (AMT) \cite{amt} over 10\% of our data across the entire skills.
The human evaluation criteria are divided into two main groups: 
1) Modular-based. The core blocks in each metric are evaluated.
2) End-to-End based. Using score-based evaluation.

\noindent --\textbf{Modular-based.}
The UniDet \cite{zhou2022simple} is the core block for counting, visual-text, spatial, color, and size composition.
First, we ask humans to visually inspect its performance and report the true positives, false positives, and false positives.
Then, we measure Person-correlation between the human F1-score and our F1-score, which shows a high correlation between our calculations and human evaluation, i.e., 93\%.

Similarly, we measure the detection and recognition accuracy of Textsnake \cite{long2018textsnake} and SAR \cite{li2019show}, respectively. 
Again, the correlation is in our favor, 98\% and 96\%, respectively.
Moreover, regarding the emotion, the annotators are asked to binary classify the emotions, i.e., positive or negative, given only the images.
The results are highly aligned with our measure, where both agree that the models generate natural images, indicating that the prompt's emotion indicator is ignored.
Regarding consistency and typos, the core module is the augmenter \cite{prithivida2021parrot, dhole2021nlaugmenter}.
To further ensure that all the generated prompts have the same meaning as the original prompt, we conduct a human study where we ask the users to rate the prompt pair (original and augmented) on a scale of 1 (not similar at all) to 5 (precisely same meaning).
The results show a great alignment, i.e., 94\%.

\noindent --\textbf{End-to-End based.}
To assess the creativity metric, we equally select 100 images per model for each hardness level and let annotators score from 1 to 5 for each generated image considering: (1) whether the generated image is creative; (2) whether the image is aligned with the given prompt.
For action composition skills, annotators are requested to assign a score between 1 to 5 based on the accuracy of the generated subject and actions in response to text prompts. 

\section{Conclusion}
\label{sec_concolusion}

We introduce a comprehensive and reliable benchmark, dubbed {\papernameAbbrev}, for evaluating text-to-image (T2I) models. 
Our benchmark measures 13 skills, categorized into five major categories, and covers 50 applications, providing a holistic evaluation of T2I models. Through our evaluation of nine recent large-scale T2I models, we have identified areas where state-of-the-art models struggle to tackle these skills, highlighting the need for continued research and development. 
Our human evaluation results confirm the effectiveness and reliability of our benchmark. 
Further, our benchmark will help ease future T2I research and progress on improving the skills covered in this benchmark.

\clearpage

{\small
\bibliographystyle{ieee_fullname}
\bibliography{egbib}
}

\appendix
\fancypagestyle{onecolumn}{%
  \fancyhf{}%
  \renewcommand{\headrulewidth}{0pt}%
  \renewcommand{\footrulewidth}{0pt}%
  \fancyfoot[C]{\thepage}%
  \onecolumn 
} 

\newpage 
\pagestyle{onecolumn} 

\section{Appendix}
This appendix includes
\begin{itemize}
    \item Qualitative results (Sec.~\ref{sec_Qualitative_results}).
    \item Prompt generation samples (Sec.~\ref{sec_Prompt_generation_samples}).
    \item Quantitative results (Sec.~\ref{sec_Quantitative_results}).
\end{itemize}

\subsection{Qualitative results}
\label{sec_Qualitative_results}

\begin{figure*}[h]
\centering
\includegraphics[width=0.99\linewidth]{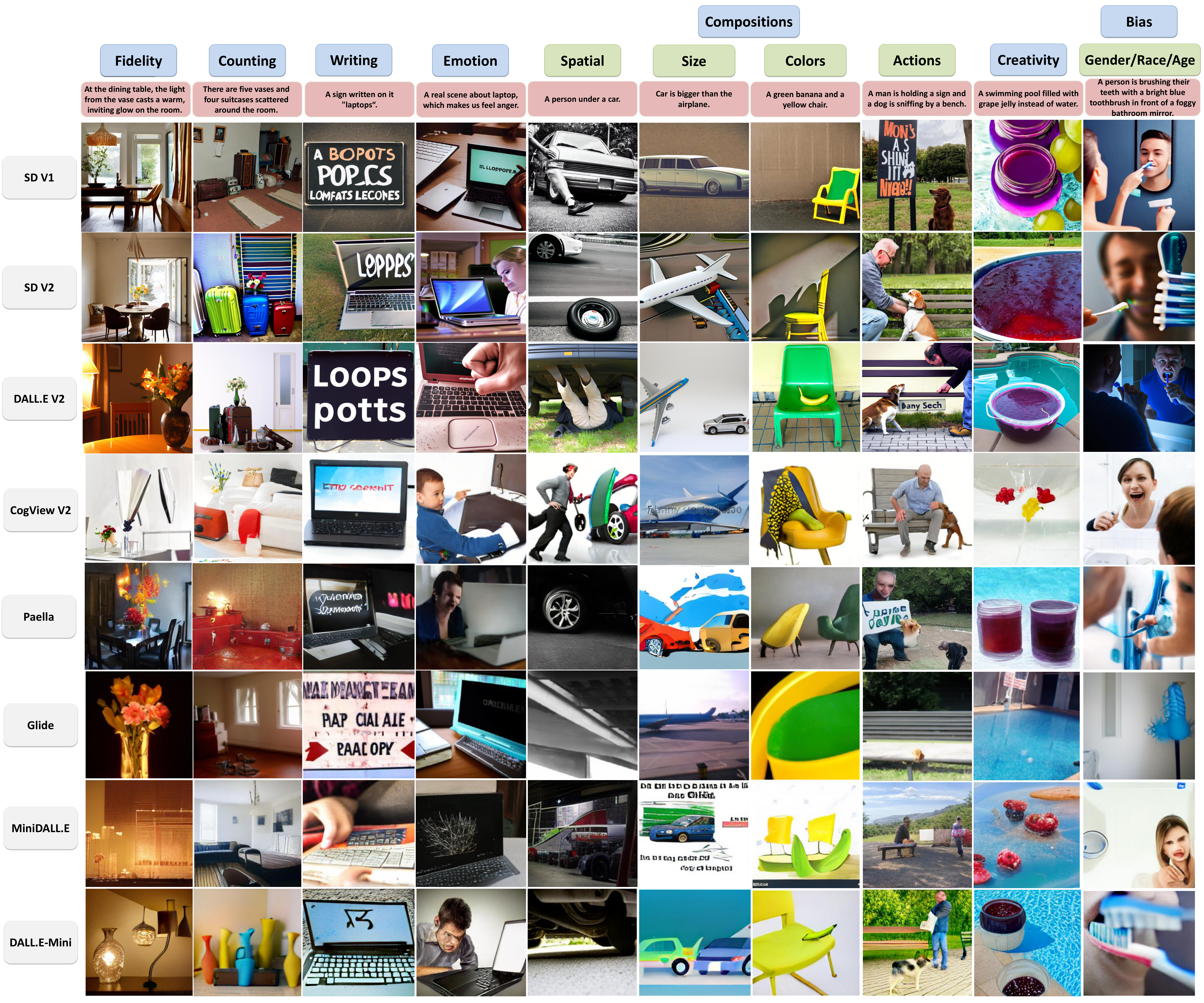}
\caption{Qualitative results. Sample \# 1.}
\label{fig_Qualitative_1}
\end{figure*}

\clearpage

\begin{figure*}
\centering
\includegraphics[width=0.99\linewidth]{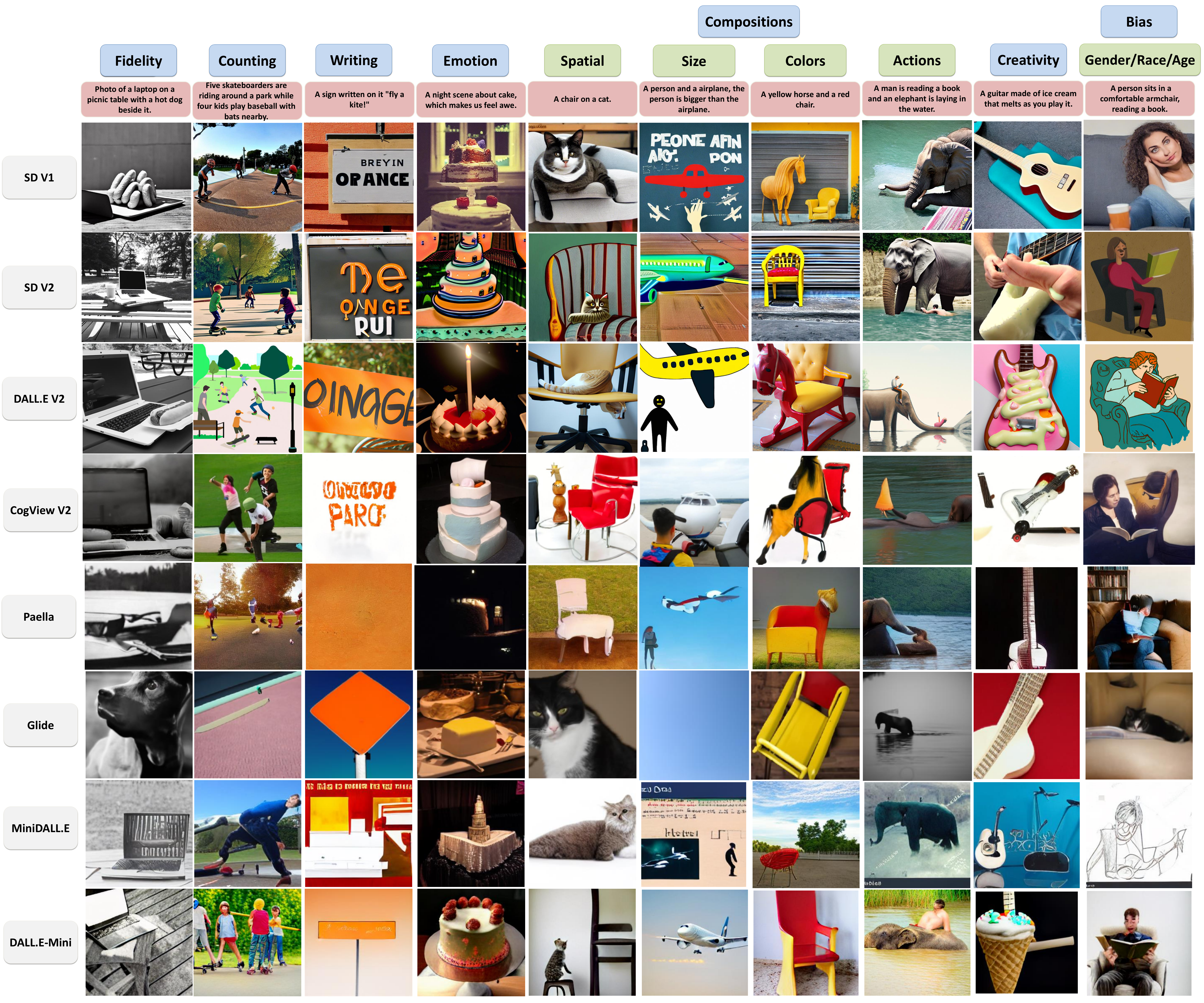}
\caption{Qualitative results. Sample \# 2.}
\label{fig_Qualitative_2}
\end{figure*}
\clearpage

\begin{figure*}
\centering
\includegraphics[width=0.99\linewidth]{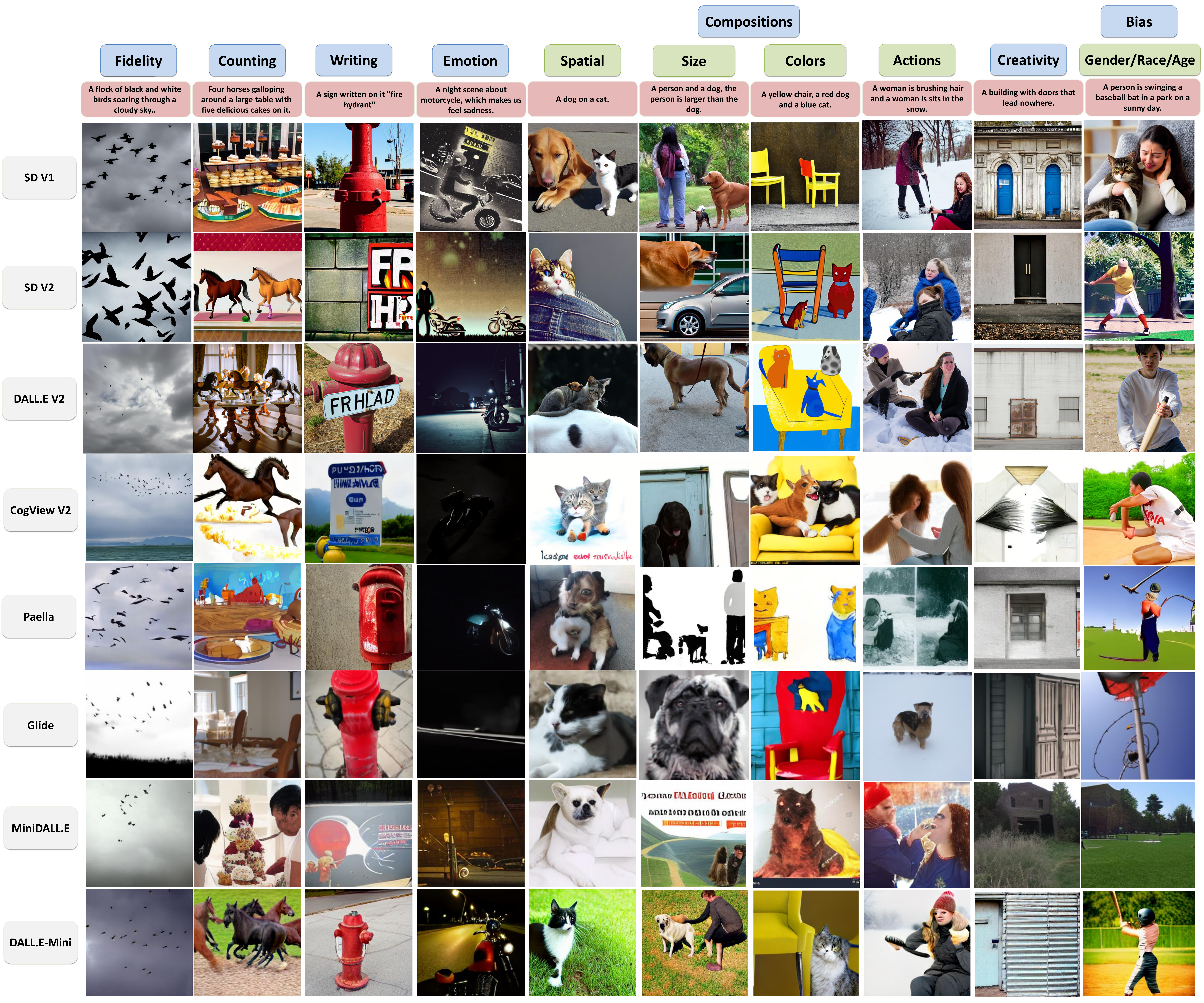}
\caption{Qualitative results. Sample \# 3.}
\label{fig_Qualitative_3}
\end{figure*}
\clearpage

\begin{figure*}
\centering
\includegraphics[width=0.99\linewidth]{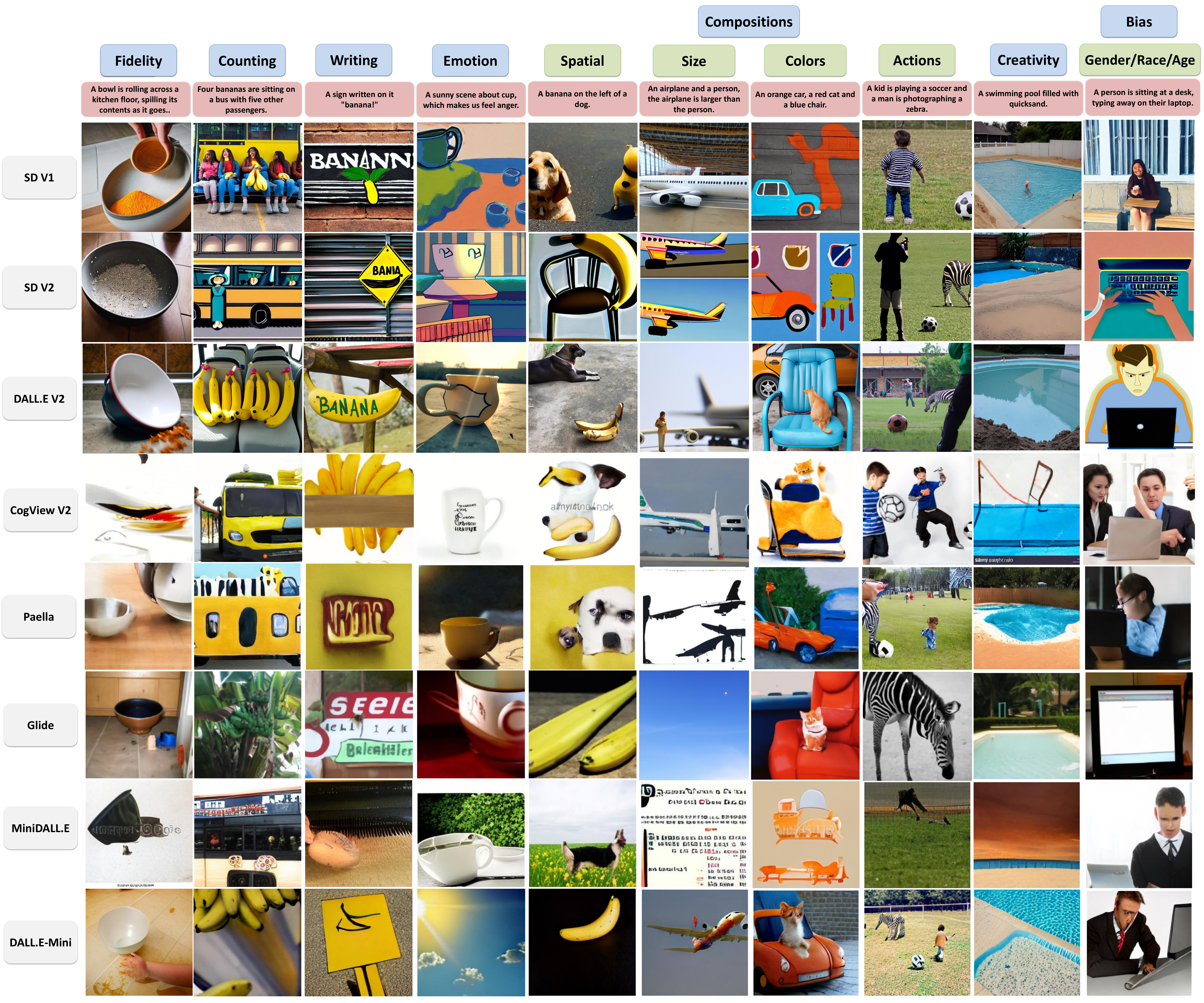}
\caption{Qualitative results. Sample \# 4.}
\label{fig_Qualitative_4}
\end{figure*}
\clearpage

\begin{figure*}
\centering
\includegraphics[width=0.99\linewidth]{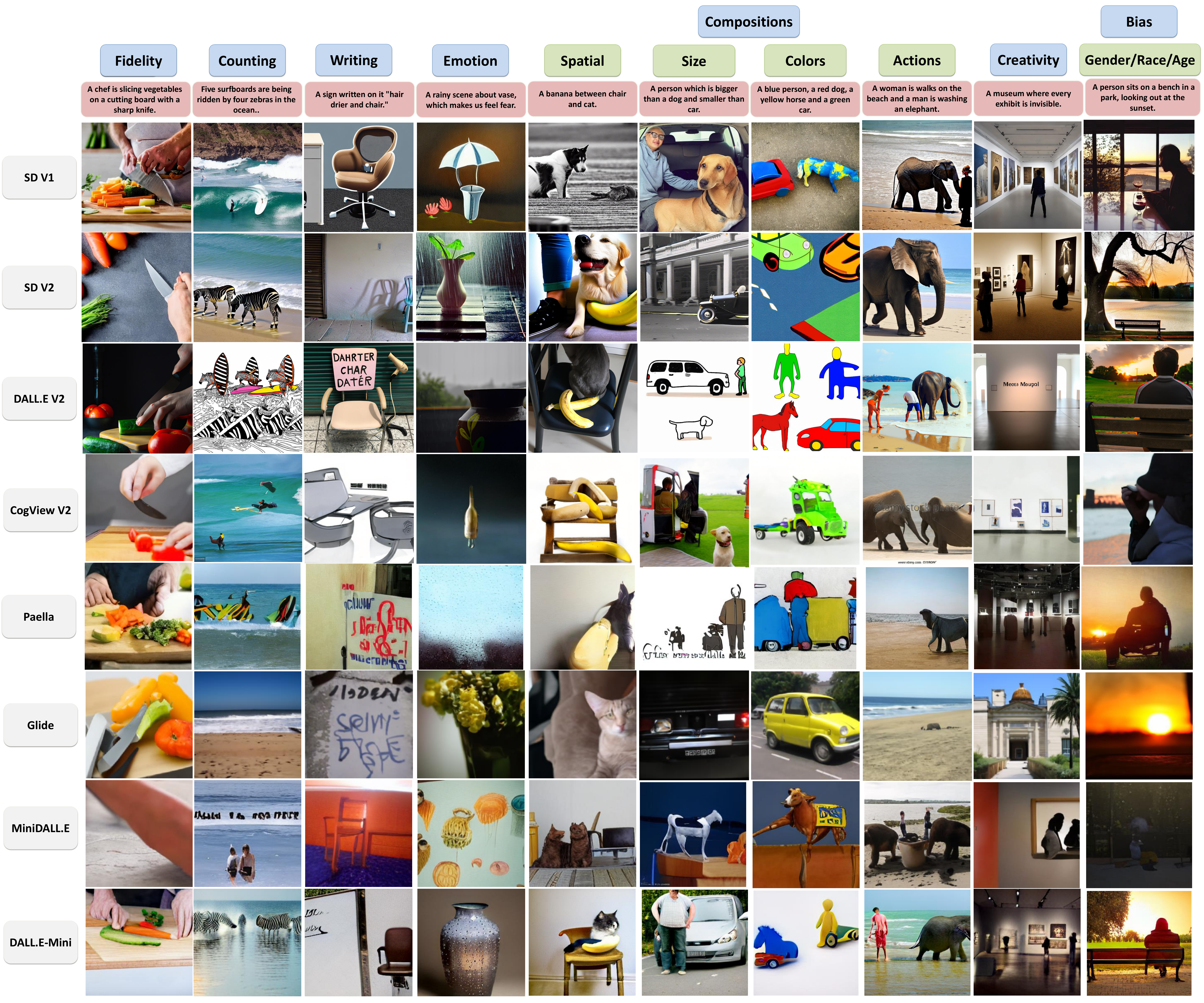}
\caption{Qualitative results. Sample \# 5.}
\label{fig_Qualitative_5}
\end{figure*}
\clearpage

\begin{figure*}
\centering
\includegraphics[width=0.99\linewidth]{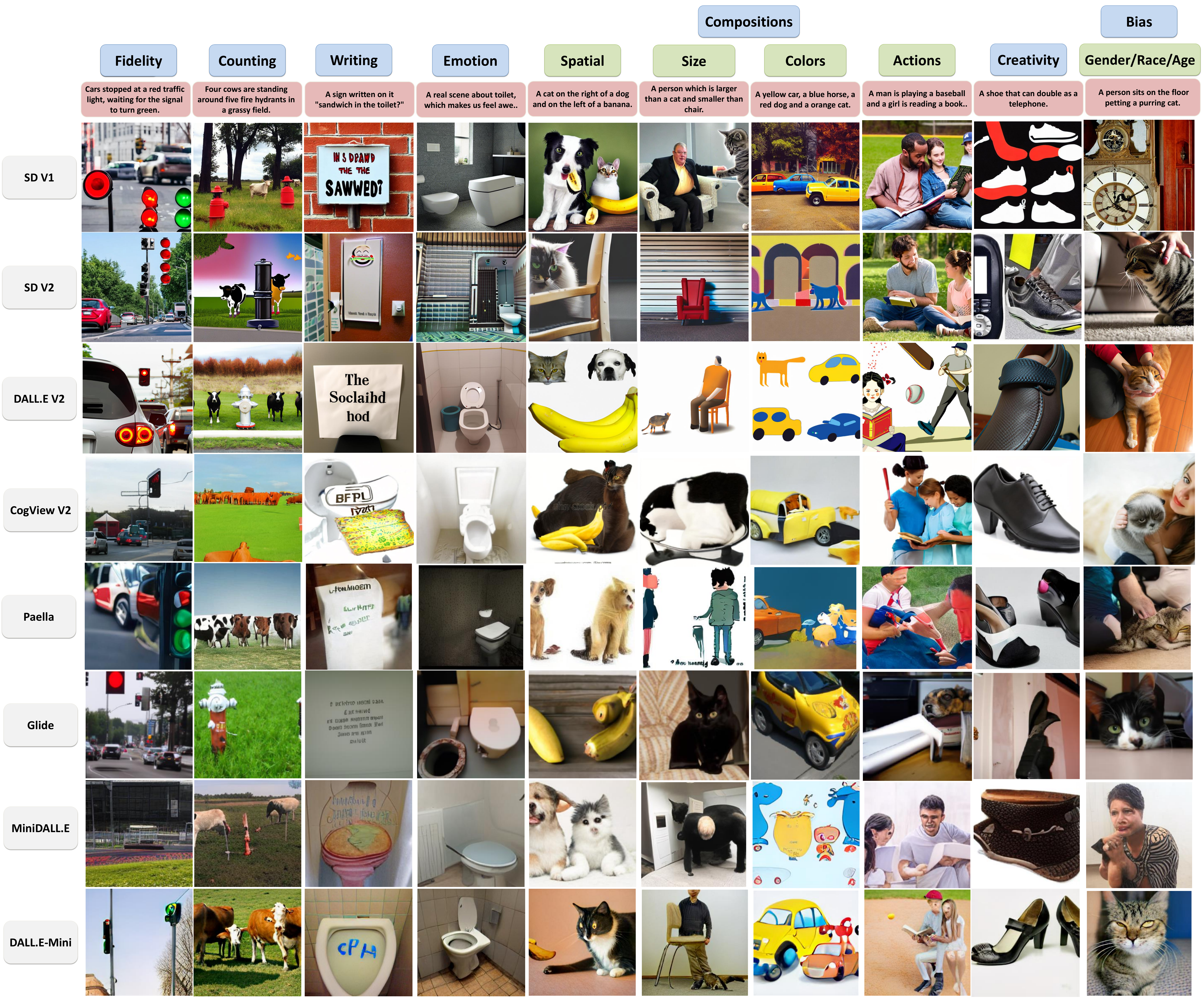}
\caption{Qualitative results. Sample \# 6.}
\label{fig_Qualitative_6}
\end{figure*}
\clearpage

\begin{figure*}
\centering
\includegraphics[width=0.99\linewidth]{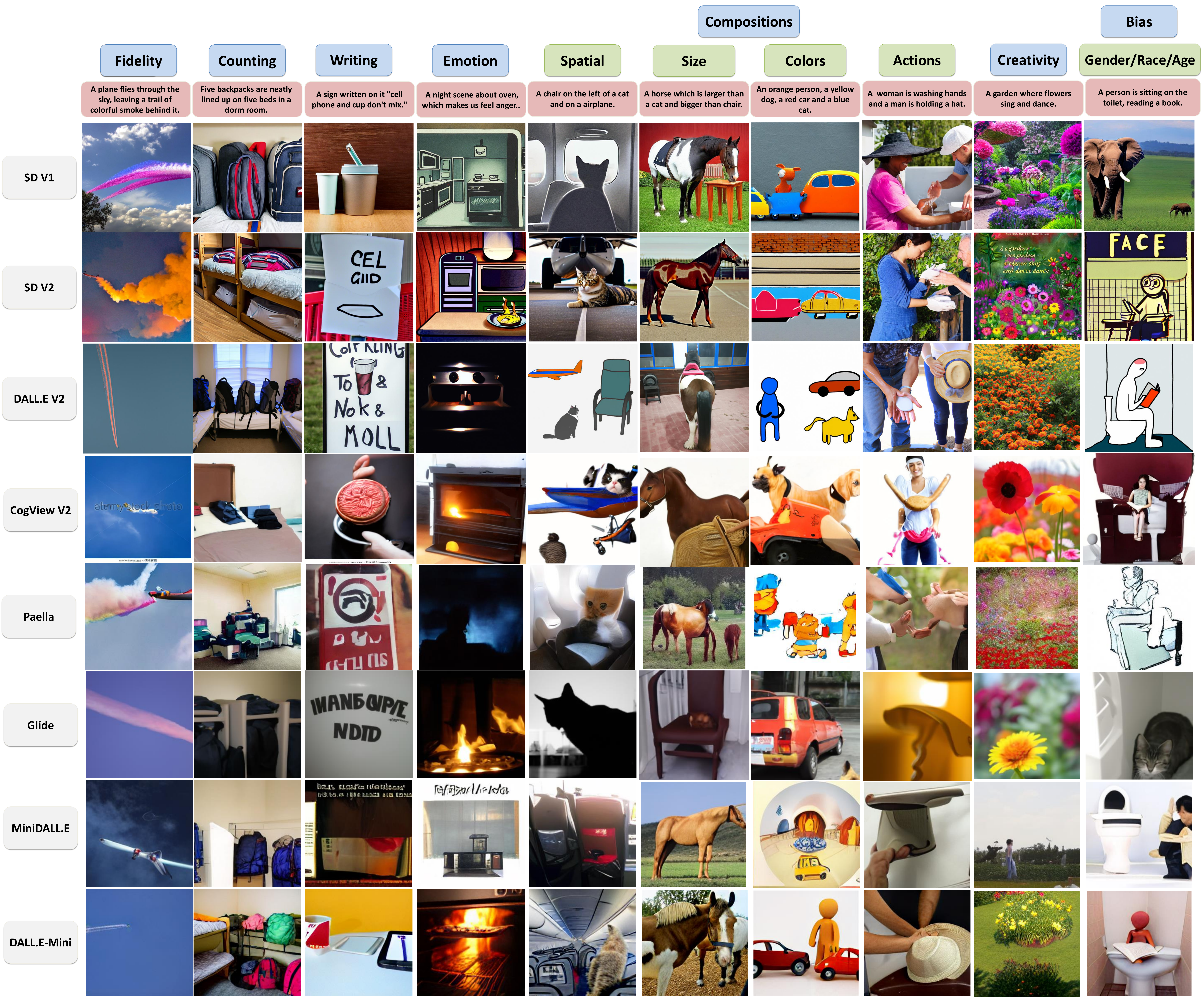}
\caption{Qualitative results. Sample \# 7.}
\label{fig_Qualitative_7}
\end{figure*}
\clearpage

\begin{figure*}
\centering
\includegraphics[width=0.99\linewidth]{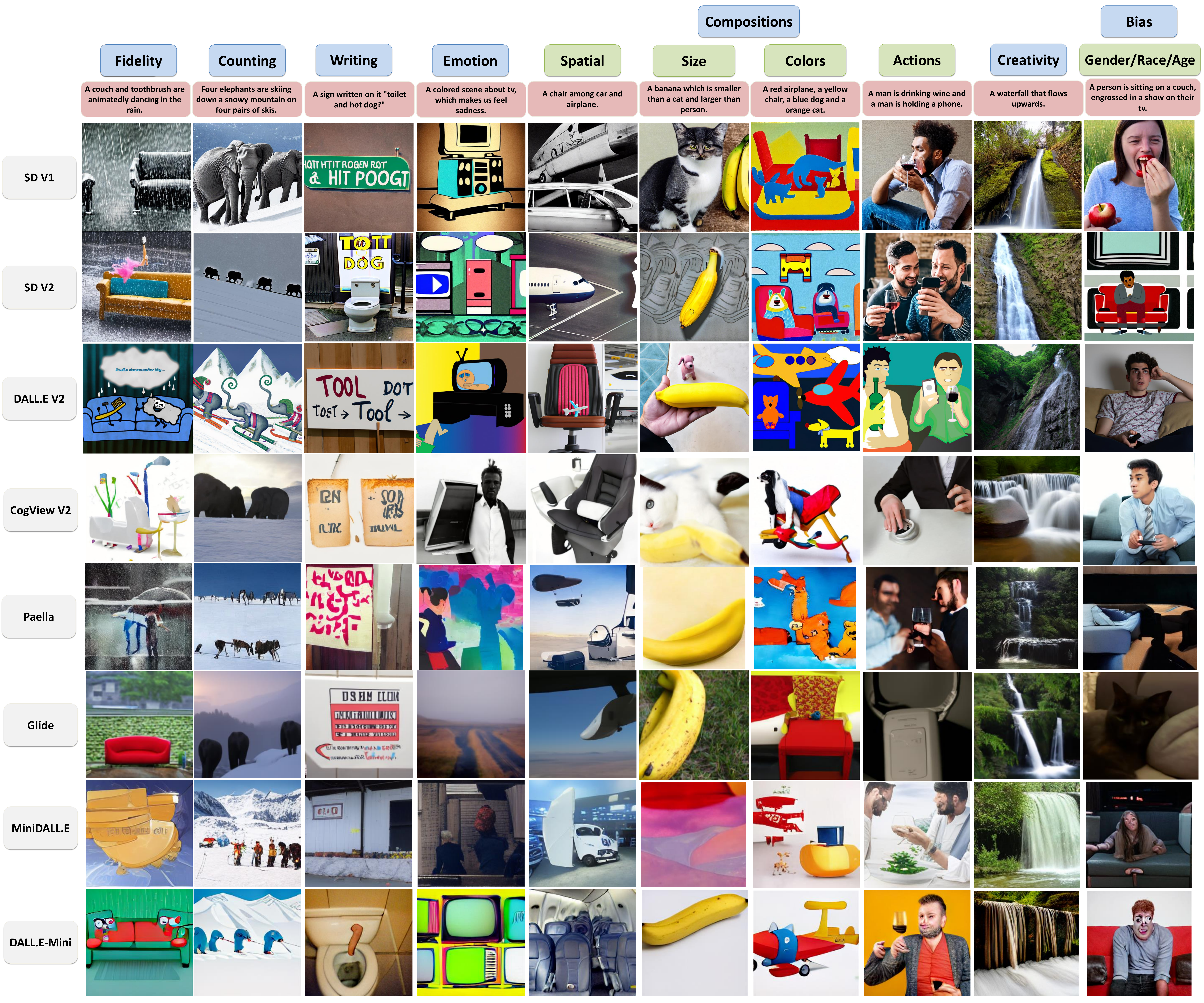}
\caption{Qualitative results. Sample \# 8.}
\label{fig_Qualitative_8}
\end{figure*}
\clearpage

\begin{figure*}
\centering
\includegraphics[width=0.99\linewidth]{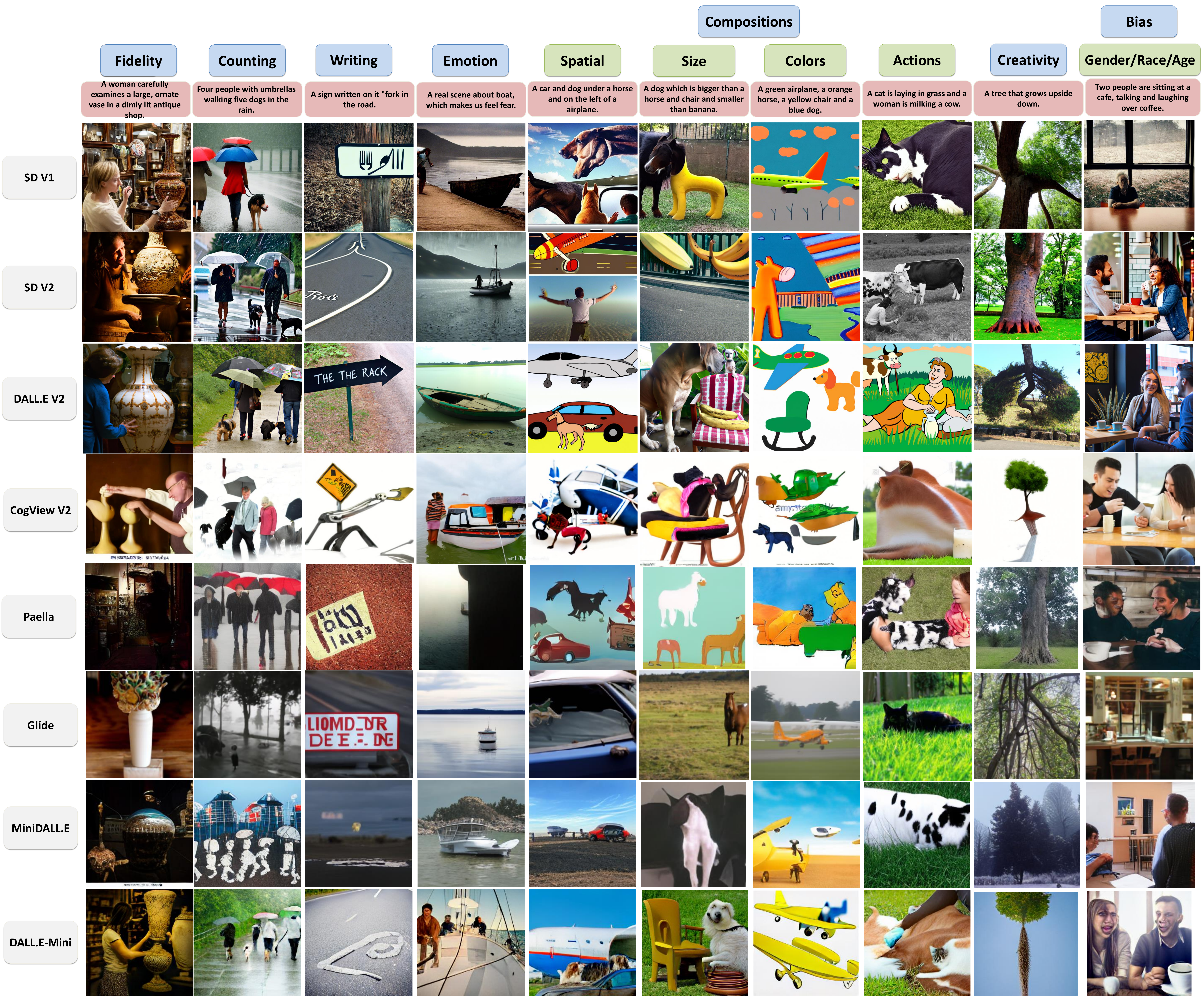}
\caption{Qualitative results. Sample \# 9.}
\label{fig_Qualitative_9}
\end{figure*}
\clearpage

\begin{figure*}
\centering
\includegraphics[width=0.99\linewidth]{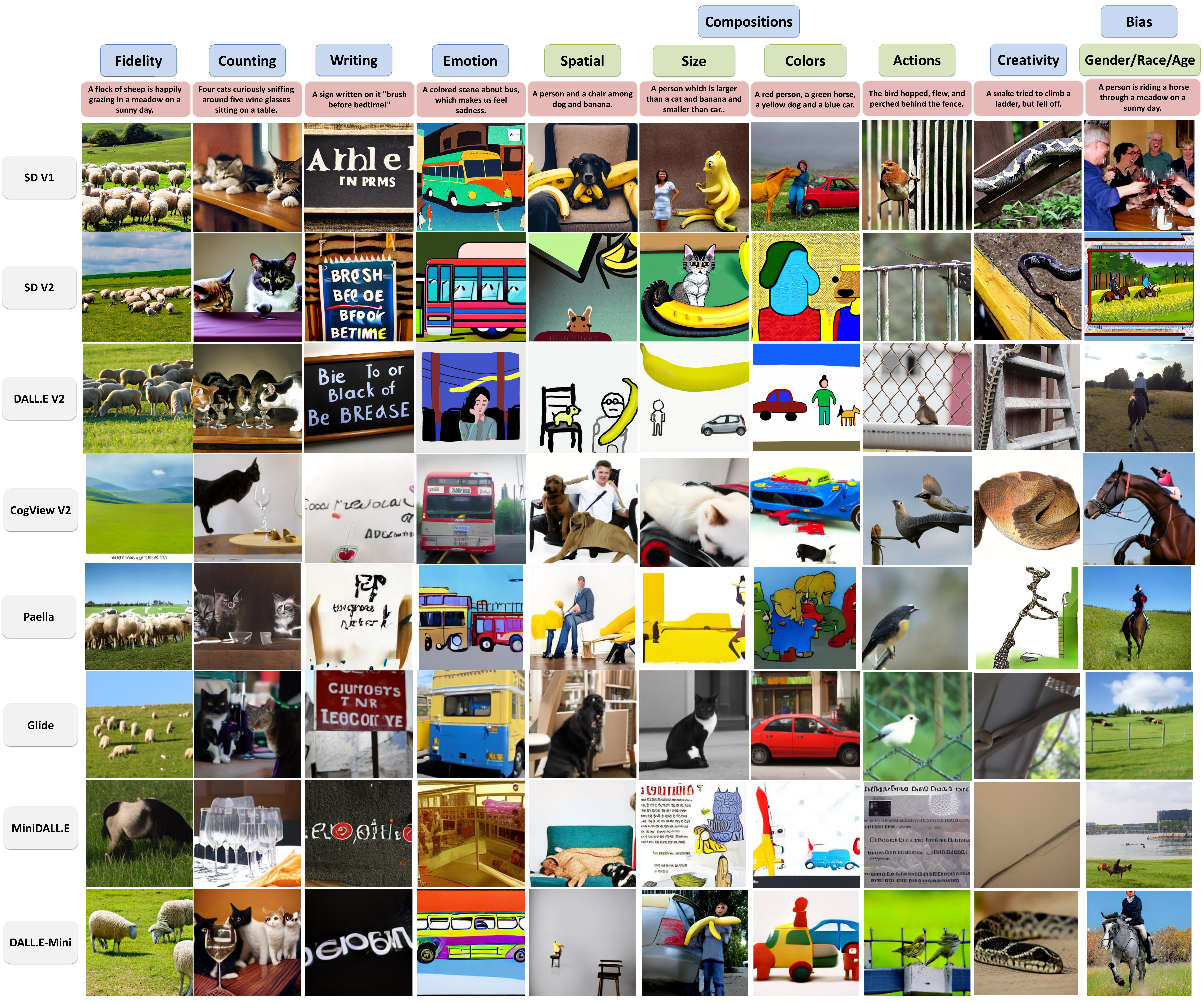}
\caption{Qualitative results. Sample \# 10.}
\label{fig_Qualitative_10}
\end{figure*}
\clearpage
\newpage

\subsection{Prompt generation samples}
\label{sec_Prompt_generation_samples}

In this section, we detail the prompt generation process for the entire evaluated skills.
Whereas, Table \ref{tab_prompt_gen_fid} shows the constrains of the fidelity prompts for each level alongside an example for each level.
Table \ref{tab_prompt_gen_count} details the rules for the counting prompts, where the number of objects are the main rule to differentiate between different levels.
Consequently, Table \ref{tab_prompt_gen_writing} depicts the visual-text template alongside ChatGPT output.
Table \ref{tab_prompt_gen_emotion} demonstrates the different templates used for each hardness level, in addition to defining the rules for each level, i.e., number of styles and objects for each level.

\begin{table*}[b]
    \centering
    \caption{Prompts generation details for fidelity skill.}
    \label{tab_prompt_gen_fid}
    \scalebox{0.80}{
    \begin{tabular}{ c c c p{0.45\linewidth} p{0.45\linewidth}}
    \toprule[1.5pt]
        Level & \# styles & \#objects & templates & examples \\
        \midrule[0.75pt]
        Easy & 1 & 1 & "Describe a \texttt{style} scene about \texttt{obj1}." & "Describe a cloudy scene about a person." \\
        
        Medium & 2 & 2 & "Describe a \texttt{style1} \texttt{style2} scene about \texttt{obj1} and \texttt{obj2}." & "Describe a sketch sunny scene about beer and car." \\
        
        Hard & 3 & 3 & "Describe a \texttt{style1} \texttt{style2} \texttt{style3} scene about \texttt{obj1}, \texttt{obj2} and \texttt{obj3}." & "Describe a black and white morning sunny scene about cake, mobile and giraffe." \\
        \bottomrule[1.5pt]
    \end{tabular}
    }
\end{table*}

\begin{table*}
    \centering
    \caption{Prompts generation details for counting skill.}
    \label{tab_prompt_gen_count}
    \scalebox{0.80}{
    \begin{tabular}{ c c c p{0.45\linewidth} p{0.45\linewidth}}
    \toprule[1.5pt]
        Level & \# styles & \# objects & templates & examples \\
        \midrule[0.75pt]
        Easy & 1 & 1 & "Describe a \texttt{style} scene about \texttt{N1} \texttt{obj1}." & "Describe a cloudy scene about a 2 tvs." \\
        
        Medium & 1 & 2 & "Describe a \texttt{style} scene about \texttt{N1} \texttt{obj1} and \texttt{N2} \texttt{obj2}." & "Describe a sunny scene about 3 beer and 2 car." \\
        
        Hard & 1 & 2 & "Describe a \texttt{style} scene about \texttt{N1} \texttt{obj1} and \texttt{N2} \texttt{obj2}." & "Describe a morning scene about 5 donuts, 4 players." \\
        \bottomrule[1.5pt]
    \end{tabular}
    }
\end{table*}

\begin{table*}
    \centering
    \caption{Prompts generation details for visual-text skill.}
    \label{tab_prompt_gen_writing}
    \scalebox{0.80}{
    \begin{tabular}{ c c p{0.45\linewidth} p{0.45\linewidth}}
    \toprule[1.5pt]
        Level & \# objects & templates & output \\
        \midrule[0.75pt]
        Easy & 1 & "\texttt{N1} words about \texttt{obj1}, the \texttt{N1} words should be between double quotes." & "laptop" \\
        
        Medium & 2 & "\texttt{N1} words about \texttt{obj1} and \texttt{obj2}, the \texttt{N1} words should be between double quotes." & "Nice vessel." \\
        
        Hard & 2 & "\texttt{N1} words about \texttt{obj1} and \texttt{obj2}, the \texttt{N1} words should be between double quotes." & "beer and cars don't make sense." \\
        \bottomrule[1.5pt]
    \end{tabular}
    }
\end{table*}

\begin{table*}
    \centering
    \caption{Prompts generation details for emotion skill.}
    \label{tab_prompt_gen_emotion}
    \scalebox{0.80}{
    \begin{tabular}{ c c c p{0.42\linewidth} p{0.42\linewidth}}
    \toprule[1.5pt]
        Level & \# styles & \# objects & templates & examples \\
        \midrule[0.75pt]
        Easy & 1 & 1 & "Describe a \texttt{style} scene about \texttt{obj1}, which makes us feel \texttt{emotion}." & "Describe a colored scene about bowl, which makes us feel contentment." \\
        Medium & 1 & 2 & "Describe a \texttt{style} scene about \texttt{obj1} and \texttt{obj2}, which makes us feel \texttt{emotion}." & "Describe a sketch scene about beer and tv, which makes us feel amusement." \\
        Hard & 2 & 2 & "Describe a \texttt{style1} \texttt{style2} scene about \texttt{obj1} and \texttt{obj2}, which makes us feel {emotion}." & "Describe a black and white morning scene about cake and giraffe, which makes us feel anger." \\
        \bottomrule[1.5pt]
    \end{tabular}
    }
\end{table*}

\begin{table*}
    \centering
    \caption{Prompts generation details for action compositions.}
    \label{tab_prompt_gen_comp_action}
    \scalebox{0.95}{
    \begin{tabular}{ c c c p{0.3\linewidth} p{0.4\linewidth}}
    \toprule[1.5pt]
        Level & Actions & Subjects & Templates & Examples \\
        \midrule[0.75pt]
        Easy & 2 & 2 & [Meta]: \texttt{text1} and \texttt{text1} & A man is feeding a dog and a cat is laying in grass. \\
        Medium & $\geq$3 & 1 & [GPT input]: Extend \texttt{text} to let the subject have at least three actions. & The man lay on the bed, stretching his arms above his head and yawning contentedly. \\
        Hard & $\geq$3 & $\geq$3 & [GPT input]: Extend \texttt{text} with other subjects doing other actions & The cat is under the bench while two children play nearby, and a woman sits nearby reading a book. \\
        \bottomrule[1.5pt]
    \end{tabular}
    }
\end{table*}

\begin{table*}
    \centering
    \caption{Prompts generation details for creativity skill.}
    \label{tab_prompt_gen_creativity}
    \scalebox{0.95}{
    \begin{tabular}{ c p{0.4\linewidth} p{0.4\linewidth}}
    \toprule[1.5pt]
        Level & Templates & Examples \\
        \midrule[0.75pt]
        easy & [Meta]: \texttt{subject} \texttt{relation} \texttt{object} (uncommon)  & The vase is in the flower. \\
        medium & [GPT input]: Describe \texttt{subject} \texttt{relation} \texttt{object} in an imaginative way that will never be seen in the real world. & The elephant is riding a person like a horse galloping through a magical forest. \\
        hard & [GPT input]: Describe a sentence for image in a counterproductive way or with personification ...  & The computer, feeling a bit lonely, asked the bottle to play a chess game. \\
        \bottomrule[1.5pt]
    \end{tabular}
    }
\end{table*}

\newpage

\subsection{Quantitative results}
\label{sec_Quantitative_results}


\begin{table}[!ht]
    \centering
    \caption{Quantitative results for counting skill across the three different hardness levels; easy, medium, and hard.}
    \scalebox{0.99}{
    \begin{tabular}{cccccccccc}
        \toprule[1.5pt]
         & \multicolumn{3}{c}{\textbf{Precision ↑}} & \multicolumn{3}{c}{\textbf{Recall ↑}} & \multicolumn{3}{c}{\textbf{F1 ↑}} \\
        \textbf{} & \textbf{Easy} & \textbf{Medium} & \textbf{Hard} & \textbf{Easy} & \textbf{Medium} & \textbf{Hard} & \textbf{Easy} & \textbf{Medium} & \textbf{Hard} \\
        \cmidrule(r){1-1}
        \cmidrule(r){2-4}
        \cmidrule(r){5-7}
        \cmidrule(r){8-10}
        \textbf{SDV1} & 67.19 & 68.66 & 75.97 & 77.76 & \textbf{43.8} & \textbf{35.71} & 72.09 & \textbf{53.48} & \textbf{48.58} \\ 
        \textbf{SDV2} & 79.79 & 84.91 & 90.81 & 67.41 & 31.58 & 25.97 & 73.07 & 46.04 & 40.39 \\ 
        \textbf{Glide} & 72.52 & 73.05 & 83.87 & 54.1 & 27.32 & 19.11 & 61.97 & 39.77 & 31.13 \\ 
        \textbf{CogView 2} & 68.32 & 67.03 & 96.47 & 63.32 & 1.22 & 0.92 & 65.73 & 2.39 & 1.82 \\ 
        \textbf{DALL.E V2} & \textbf{81.71} & 83.88 & \textbf{98.28} & \textbf{82} & 1.52 & 0.85 & \textbf{81.85} & 2.99 & 1.7 \\ 
        \textbf{Paella} & 73.93 & 70.21 & 77.66 & 69.12 & 31.27 & 23.16 & 71.44 & 43.27 & 35.68 \\ 
        \textbf{minDALL-E} & 76.89 & 79.71 & 89.05 & 48.33 & 20.98 & 14.05 & 59.35 & 33.21 & 24.27 \\ 
        \textbf{DALL-E\_Mini} & 76.98 & \textbf{86.75} & 96.66 & 78.32 & 1.22 & 0.84 & 77.63 & 2.41 & 1.67 \\ 
        \bottomrule[1.5pt]
    \end{tabular}
    }
\end{table}

\begin{table}[!ht]
    \centering
    \caption{Ablation study for the impact of the prompt details on the counting skill across the three different hardness levels; easy, medium, and hard.}
    \scalebox{0.99}{
    \begin{tabular}{cccccccccc}
        \toprule[1.5pt]
         & \multicolumn{3}{c}{\textbf{Precision ↑}} & \multicolumn{3}{c}{\textbf{Recall ↑}} & \multicolumn{3}{c}{\textbf{F1 ↑}} \\
        \textbf{} & \textbf{Vanilla} & \textbf{Meta} & \textbf{Detailed} & \textbf{Vanilla} & \textbf{Meta} & \textbf{Detailed} & \textbf{Vanilla} & \textbf{Meta} & \textbf{Detailed} \\
        \cmidrule(r){1-1}
        \cmidrule(r){2-4}
        \cmidrule(r){5-7}
        \cmidrule(r){8-10}
        \textbf{SDV1} & 75.81 & 69.14 & 70.61 & \textbf{46.48} & \textbf{46.7} & \textbf{52.42} & \textbf{55.21} & \textbf{52.79} & \textbf{58.05} \\ 
        \textbf{SDV2} & 78.27 & 75.78 & 85.17 & 36.02 & 39.25 & 41.65 & 46.04 & 46.77 & 53.16 \\ 
        \textbf{Glide} & 87.91 & 80.79 & 76.48 & 30.19 & 28.27 & 33.51 & 41.96 & 38.72 & 44.29 \\ 
        \textbf{CogView 2} & 85.22 & 85.89 & 79.11 & 19.23 & 19.8 & 21.9 & 22.3 & 22.13 & 23.03 \\ 
        \textbf{DALL.E V2} & \textbf{92.53} & \textbf{90.29} & \textbf{87.96} & 28.7 & 27.93 & 28.12 & 29.99 & 29.06 & 28.85 \\ 
        \textbf{Paella} & 80.52 & 72.53 & 73.93 & 38.26 & 39.41 & 41.19 & 49.82 & 47.98 & 50.13 \\ 
        \textbf{minDALL-E} & 86.09 & 87.7 & 81.88 & 19.51 & 16.86 & 27.78 & 29.22 & 26.59 & 38.94 \\ 
        \textbf{DALL-E\_Mini} & 89.21 & 87.19 & 86.8 & 24.32 & 23.93 & 26.8 & 26.71 & 25.94 & 27.24 \\ 
        \bottomrule[1.5pt]
    \end{tabular}
    }
\end{table}

\begin{table}[!ht]
    \centering
    \caption{Quantitative results for emotion skill.}
    \scalebox{0.95}{
    \begin{tabular}{ccccccccccc}
        \toprule[1.5pt]
        \textbf{} & \multicolumn{4}{c}{\textbf{K=5}} & \multicolumn{4}{c}{\textbf{K=10}} & \textbf{} & \textbf{} \\
        \textbf{} & \textbf{ClipScore} & \textbf{CIDEr} & \textbf{BLEU-1} & \textbf{BLEU-4} & \textbf{ClipScore} & \textbf{CIDEr} & \textbf{BLEU-1} & \textbf{BLEU-4} & \textbf{CLS 8 classes} & \textbf{CLS 2 classes} \\ 
        \cmidrule(r){1-1}
        \cmidrule(r){2-5}
        \cmidrule(r){6-9}
        \cmidrule(r){10-10}
        \cmidrule(r){11-11}
        \textbf{SDV1} & 0.33 & 0.80 & 0.24 & 0.09 & 0.34 & 0.91 & 0.26 & 0.10 & 0.14 & 0.54 \\ 
        \textbf{SDV2} & 0.32 & 0.77 & 0.23 & 0.09 & 0.32 & 0.88 & 0.25 & 0.10 & 0.15 & 0.53 \\ 
        \textbf{Glide} & 0.30 & 0.73 & 0.22 & 0.08 & 0.30 & 0.82 & 0.24 & 0.09 & 0.14 & 0.52 \\ 
        \textbf{CogView 2} & 0.30 & 0.71 & 0.22 & 0.08 & 0.31 & 0.81 & 0.24 & 0.09 & 0.16 & 0.53 \\ 
        \textbf{DALL.E V2} & 0.35 & 0.88 & 0.26 & 0.10 & 0.35 & 1.00 & 0.28 & 0.11 & 0.13 & 0.50 \\ 
        \textbf{Paella} & 0.32 & 0.73 & 0.22 & 0.08 & 0.32 & 0.82 & 0.24 & 0.09 & 0.14 & 0.52 \\ 
        \textbf{minDALL-E} & 0.28 & 0.65 & 0.21 & 0.07 & 0.28 & 0.75 & 0.22 & 0.08 & 0.15 & 0.52 \\ 
        \textbf{DALL-E\_Mini} & 0.33 & 0.73 & 0.23 & 0.09 & 0.34 & 0.85 & 0.25 & 0.10 & 0.16 & 0.55 \\ 
        \bottomrule[1.5pt]
    \end{tabular}
    }
\end{table}

\begin{table*}
\begin{minipage}[c]{0.30\textwidth}
    \captionof{table}{Quantitative results for visual text skill.}
    \label{tab_visual_text}
    \scalebox{0.81}{
    \begin{tabular}{ccc}
        \toprule[1.5pt]
        \textbf{} & \textbf{NED ↓} & \textbf{CER ↓} \\
        \cmidrule(r){1-1}
        \cmidrule(r){2-3}
        \textbf{SDV1} & 84.98 & 92.27 \\ 
        \textbf{SDV2} & 83.16 & 94.52 \\ 
        \textbf{Glide} & 89.92 & 95.25 \\ 
        \textbf{CogView 2} & 89.55 & 96.87 \\ 
        \textbf{DALL.E V2} & \textbf{74.89} & \textbf{87.46} \\ 
        \textbf{Paella} & 89.83 & 97.37 \\ 
        \textbf{minDALL-E} & 90.85 & 96.44 \\ 
        \textbf{DALL-E\_Mini} & 94.06 & 99.42 \\
        \bottomrule[1.5pt]
    \end{tabular}
    }
\end{minipage}
\hfill
\begin{minipage}[c]{0.33\textwidth}
    \captionof{table}{Quantitative results for consistency skill.}
    \label{tab_consistency}
    \scalebox{0.80}{
    \begin{tabular}{cccc}
        \toprule[1.5pt]
        \textbf{} & \textbf{Easy} & \textbf{Medium} & \textbf{Hard} \\
        \cmidrule(r){1-1}
        \cmidrule(r){2-4}
        \textbf{SDV1} & 0.79 & 0.79 & 0.78 \\
        \textbf{SDV2} & 0.81 & 0.80 & 0.80 \\
        \textbf{Glide} & 0.78 & 0.78 & 0.77 \\
        \textbf{CogView 2} & 0.72 & 0.71 & 0.71 \\
        \textbf{DALL.E V2} & 0.82 & 0.81 & 0.80 \\
        \textbf{Paella} & 0.82 & 0.81 & 0.81 \\
        \textbf{minDALL-E} & 0.72 & 0.72 & 0.71 \\ 
        \textbf{DALL-E\_Mini} & 0.82 & 0.81 & 0.81 \\ 
        \bottomrule[1.5pt]
    \end{tabular}
    }
\end{minipage}
\hfill
\begin{minipage}[c]{0.33\textwidth}
    \captionof{table}{Quantitative results for typos skill.}
    \label{tab_typos}
    \scalebox{0.75}{
    \begin{tabular}{cccc}
        \toprule[1.5pt]
        \textbf{} & \textbf{Easy} & \textbf{Medium} & \textbf{Hard} \\
        \cmidrule(r){1-1}
        \cmidrule(r){2-4}
        \textbf{SDV1} & 0.78 & 0.76 & 0.73 \\
        \textbf{SDV2} & 0.80 & 0.77 & 0.73 \\
        \textbf{Glide} & 0.77 & 0.74 & 0.74 \\
        \textbf{CogView 2} & 0.71 & 0.70 & 0.68 \\
        \textbf{DALL.E V2} & 0.81 & 0.80 & 0.78 \\
        \textbf{Paella} & 0.81 & 0.79 & 0.77 \\
        \textbf{minDALL-E} & 0.72 & 0.70 & 0.69 \\
        \textbf{DALL-E\_Mini} & 0.80 & 0.77 & 0.74 \\ 
        \bottomrule[1.5pt]
    \end{tabular}
    }
\end{minipage}
\end{table*}

\begin{table*}
\begin{minipage}[c]{0.30\textwidth}
    \centering
    \captionof{table}{Quantitative results for gender bias.}
    \label{tab_visual_text}
    \scalebox{0.99}{
    \begin{tabular}{cc}
        \toprule[1.5pt]
        \textbf{Bias} & \textbf{MAD \%} \\
        \cmidrule(r){1-1}
        \cmidrule(r){2-2}
        \textbf{SDV1} & 7.94 \\ 
        \textbf{SDV2} & 18.51 \\ 
        \textbf{CogView 2} & 17.83 \\ 
        \textbf{DALL.E V2} & 18.05 \\ 
        \textbf{minDALL-E} & 23.07 \\ 
        \bottomrule[1.5pt]
    \end{tabular}
    }
\end{minipage}
\hfill
\begin{minipage}[c]{0.65\textwidth}
    \centering
    \captionof{table}{Quantitative results for fairness skill.}
    \label{tab_consistency}
    \scalebox{0.90}{
    \begin{tabular}{ccc}
        \toprule[1.5pt]
        \textbf{} & \textbf{Fairness Score Gender} & \textbf{Fairness Score Styles} \\ 
        \cmidrule(r){1-1}
        \cmidrule(r){2-3}
        \textbf{SDV1} & 1.41 & 0.10 \\ 
        \textbf{SDV2} & 0.63 & 0.11 \\ 
        \textbf{Glide} & 0.36 & 0.06 \\ 
        \textbf{CogView 2} & 3.42 & 0.06 \\ 
        \textbf{DALL.E V2} & 1.71 & 0.11 \\ 
        \textbf{Paella} & 1.90 & 0.09 \\ 
        \textbf{minDALL-E} & 0.51 & 0.11 \\ 
        \textbf{DALL-E\_Mini} & 1.67 & 0.11 \\
        \bottomrule[1.5pt]
    \end{tabular}
    }
\end{minipage}
\end{table*}

\begin{table}[!ht]
    \centering
    \caption{Quantitative results for spatial, size, and colors composition skills.}
    \scalebox{0.99}{
    \begin{tabular}{cccccccccc}
        \toprule[1.5pt]
        \textbf{} & \multicolumn{3}{c}{\textbf{Spatial ↑}} & \multicolumn{3}{c}{\textbf{Size ↑}} & \multicolumn{3}{c}{\textbf{Colors ↑}} \\
        \textbf{} & \textbf{Easy} & \textbf{Medium} & \textbf{Hard} & \textbf{Easy} & \textbf{Medium} & \textbf{Hard} & \textbf{Easy} & \textbf{Medium} & \textbf{Hard} \\ 
        \cmidrule(r){1-1}
        \cmidrule(r){2-4}
        \cmidrule(r){5-7}
        \cmidrule(r){8-10}
        \textbf{SDV1} & 21.75 & 0 & 0 & 27.34 & 0 & 0 & 30 & 0 & 0 \\ 
        \textbf{SDV2} & 1.19 & 0 & 0 & 0.19 & 0.19 & 0 & 20 & 0 & 0 \\ 
        \textbf{Glide} & 2.49 & 0 & 0 & 6.78 & 0 & 0 & 15 & 0 & 0 \\ 
        \textbf{CogView 2} & 8.88 & 0 & 0 & 11.97 & 0 & 0 & 15 & 0 & 0 \\ 
        \textbf{DALL.E V2} & 28.34 & 0 & 0 & 29.94 & 0 & 0 & 38 & 0 & 0 \\ 
        \textbf{Paella} & 8.78 & 0 & 0 & 7.38 & 0 & 0 & 3 & 0 & 0 \\ 
        \textbf{minDALL-E} & 4.29 & 0 & 0 & 2.19 & 0 & 0 & 2 & 0 & 0 \\
        \textbf{DALL-EMini} & 15.17 & 0 & 0 & 19.16 & 0 & 0 & 35 & 0 & 0 \\
        \bottomrule[1.5pt]
    \end{tabular}
    }
\end{table}

\begin{table}[!ht]
    \centering
    \caption{Quantitative results for actions composition skill.}
    \scalebox{0.70}{
    \begin{tabular}{cccccccccccccccc}
        \toprule[1.5pt]
        \textbf{} & \multicolumn{5}{c}{\textbf{Easy}} & \multicolumn{5}{c}{\textbf{Medium}} & \multicolumn{5}{c}{\textbf{Hard}} \\ 
        \textbf{} & \textbf{BLEU1} & \textbf{BLEU2} & \textbf{BLEU3} & \textbf{BLEU4} & \textbf{CIDEr} & \textbf{BLEU1} & \textbf{BLEU2} & \textbf{BLEU3} & \textbf{BLEU4} & \textbf{CIDEr} & \textbf{BLEU1} & \textbf{BLEU2} & \textbf{BLEU3} & \textbf{BLEU4} & \textbf{CIDEr} \\
        \cmidrule(r){1-1}
        \cmidrule(r){2-6}
        \cmidrule(r){7-11}
        \cmidrule(r){12-16}
        \textbf{SDV1} & 0.57 & 0.47 & 0.37 & 0.29 & 2.40 & 0.35 & 0.25 & 0.18 & 0.14 & 1.14 & 0.36 & 0.27 & 0.20 & 0.15 & 0.64 \\ 
        \textbf{SDV2} & 0.57 & 0.47 & 0.37 & 0.29 & 2.32 & 0.37 & 0.27 & 0.19 & 0.14 & 1.14 & 0.37 & 0.27 & 0.20 & 0.15 & 0.69 \\ 
        \textbf{Glide} & 0.46 & 0.34 & 0.25 & 0.19 & 1.69 & 0.29 & 0.19 & 0.13 & 0.09 & 0.88 & 0.28 & 0.19 & 0.14 & 0.11 & 0.51 \\ 
        \textbf{CogView 2} & 0.53 & 0.43 & 0.33 & 0.25 & 2.10 & 0.33 & 0.23 & 0.16 & 0.12 & 1.0004 & 0.33 & 0.24 & 0.17 & 0.13 & 0.63 \\ 
        \textbf{DALL.E V2} & 0.63 & 0.54 & 0.43 & 0.34 & 2.46 & 0.33 & 0.23 & 0.16 & 0.12 & 1.16 & 0.39 & 0.29 & 0.21 & 0.16 & 0.73 \\ 
        \textbf{Paella} & 0.51 & 0.41 & 0.31 & 0.23 & 1.93 & 0.33 & 0.23 & 0.17 & 0.13 & 1.03 & 0.32 & 0.22 & 0.16 & 0.12 & 0.56 \\ 
        \textbf{minDALL-E} & 0.49 & 0.38 & 0.28 & 0.21 & 1.82 & 0.31 & 0.21 & 0.15 & 0.11 & 0.90 & 0.31 & 0.21 & 0.15 & 0.11 & 0.57 \\ 
        \bottomrule[1.5pt]
    \end{tabular}
    }
\end{table}

\begin{table}[!ht]
    \centering
    \caption{Quantitative results for creativity skill.}
    \scalebox{0.57}{
    \begin{tabular}{ccccccccccccccccccc}
        \toprule[1.5pt]
        \textbf{} & \multicolumn{6}{c}{\textbf{Easy}} & \multicolumn{6}{c}{\textbf{Medium}} & \multicolumn{6}{c}{\textbf{Hard}} \\ 
        \textbf{} & \textbf{deviation} & \textbf{BLEU1} & \textbf{BLEU2} & \textbf{BLEU3} & \textbf{BLEU4} & \textbf{CIDEr} & \textbf{deviation} & \textbf{BLEU1} & \textbf{BLEU2} & \textbf{BLEU3} & \textbf{BLEU4} & \textbf{CIDEr} & \textbf{deviation} & \textbf{BLEU1} & \textbf{BLEU2} & \textbf{BLEU3} & \textbf{BLEU4} & \textbf{CIDEr} \\
        \cmidrule(r){1-1}
        \cmidrule(r){2-7}
        \cmidrule(r){8-13}
        \cmidrule(r){14-19}
        \textbf{SDV1} & 0.34 & 0.42 & 0.30 & 0.22 & 0.16 & 0.64 & 0.32 & 0.40 & 0.29 & 0.21 & 0.16 & 0.65 & 0.34 & 0.32 & 0.21 & 0.14 & 0.11 & 0.35 \\ 
        \textbf{SDV2} & 0.34 & 0.42 & 0.31 & 0.22 & 0.17 & 0.66 & 0.33 & 0.42 & 0.31 & 0.22 & 0.17 & 0.66 & 0.35 & 0.33 & 0.23 & 0.16 & 0.12 & 0.36 \\ 
        \textbf{Glide} & 0.29 & 0.38 & 0.27 & 0.19 & 0.14 & 0.57 & 0.29 & 0.37 & 0.25 & 0.17 & 0.14 & 0.56 & 0.29 & 0.29 & 0.19 & 0.13 & 0.10 & 0.30 \\ 
        \textbf{CogView 2} & 0.33 & 0.38 & 0.26 & 0.19 & 0.14 & 0.56 & 0.30 & 0.38 & 0.26 & 0.18 & 0.14 & 0.56 & 0.28 & 0.28 & 0.18 & 0.13 & 0.10 & 0.27 \\ 
        \textbf{DALL.E V2} & 0.29 & 0.43 & 0.32 & 0.23 & 0.17 & 0.71 & 0.30 & 0.44 & 0.32 & 0.24 & 0.18 & 0.68 & 0.28 & 0.33 & 0.23 & 0.16 & 0.12 & 0.37 \\ 
        \textbf{Paella} & 0.29 & 0.40 & 0.28 & 0.20 & 0.15 & 0.59 & 0.28 & 0.40 & 0.28 & 0.20 & 0.15 & 0.60 & 0.29 & 0.31 & 0.20 & 0.14 & 0.10 & 0.31 \\ 
        \textbf{minDALL-E} & 0.33 & 0.37 & 0.25 & 0.17 & 0.13 & 0.52 & 0.32 & 0.35 & 0.23 & 0.16 & 0.12 & 0.51 & 0.33 & 0.26 & 0.16 & 0.11 & 0.08 & 0.24 \\ 
        \textbf{DALL-E\_Mini} & 0.31 & 0.42 & 0.30 & 0.22 & 0.16 & 0.63 & 0.29 & 0.41 & 0.30 & 0.21 & 0.1639 & 0.64 & 0.29 & 0.32 & 0.22 & 0.15 & 0.11 & 0.36 \\ 
        \bottomrule[1.5pt]
    \end{tabular}
    }
\end{table}

\end{document}